%% file: main.tex
\documentclass[10pt,twocolumn,letterpaper]{article}

\usepackage[pagenumbers]{cvpr} 

\input{preamble}

\definecolor{cvprblue}{rgb}{0.21,0.49,0.74}
\usepackage[pagebackref,breaklinks,colorlinks,citecolor=cvprblue]{hyperref}

\usepackage[utf8]{inputenc} 
\usepackage[T1]{fontenc}    
\usepackage{url}            
\usepackage{booktabs}       
\usepackage{amsfonts}       
\usepackage{nicefrac}       
\usepackage{microtype}      
\usepackage{xcolor}         
\definecolor{ball blue}{rgb}{0.13, 0.67, 0.8}
\definecolor{bleudefrance}{rgb}{0.19, 0.55, 0.91}
\usepackage{bbding}
\usepackage{pifont}
\usepackage{wasysym}
\usepackage{amssymb}

\usepackage{amsmath}
\usepackage{graphicx}
\usepackage{multirow}
\allowdisplaybreaks[4]
\usepackage{amssymb}
\usepackage{array}
\usepackage{enumitem}
\usepackage{color}
\usepackage{colortbl}
\usepackage{bm}
\usepackage{amsmath}
\usepackage{float}
\definecolor{ForestGreen}{rgb}{0.13, 0.55, 0.13}
\newcommand{\stdvu}[1]{\tiny{\color{darkgray}(#1)} {\color{ForestGreen}$\uparrow$}}

\definecolor{mypink}{rgb}{.99,.91,.95}

\usepackage{wrapfig}

\usepackage{bbding}
\usepackage{pifont}
\usepackage{wasysym}
\usepackage{amssymb}
\usepackage[export]{adjustbox}

\definecolor{firebrick}{rgb}{0.7, 0.13, 0.13}
\definecolor{darkpastelgreen}{rgb}{0.01, 0.75, 0.24}
\definecolor{deepskyblue}{rgb}{0.0, 0.75, 1.0}
\definecolor{mypink2}{rgb}{.99,.96,.98}
\definecolor{mypink1}{rgb}{.99,.93,.98}
\definecolor{mypink}{rgb}{.99,.90,.98}
\definecolor{mygray}{rgb}{.95,.95,.95}
\definecolor{lv14}{rgb}{0.5,0.5,0.5}

\definecolor{myblue}{rgb}{0.94, 0.95, 1.0}

\def\paperID{2598} 
\def\confName{CVPR}
\def\confYear{2024}

\title{VQA4CIR: Boosting Composed Image Retrieval with Visual Question Answering}

\author{Chun-Mei Feng$^1$\quad Yang Bai$^1$\thanks{Corresponding author.}\quad Tao Luo$^1$\quad Zhen Li$^2$\quad Salman Khan$^{3,4}$\\
Wangmeng Zuo$^5$\quad Xinxing Xu$^1$\quad Rick Siow Mong Goh$^1$\quad Yong Liu$^1$\\
$^1$IHPC, A*STAR, Singapore;
$^2$CUHK, Shenzhen, China;\\
$^3$MBZUAI, UAE;
$^4$ANU, Australia;
$^5$HIT, Harbin, China\\
{\tt\small fengcm.ai@gmail.com; baiyangolf@gmail.com}
\\ \small {\url{https://github.com/chunmeifeng/VQA4CIR}}
}

\begin{document}
\maketitle
\input{sec/0_abstract}

\input{sec/1_Intro}

\input{sec/2_RelatedWork}

\input{sec/3_Methodology}

\input{sec/4_Experiments}

\input{sec/5_Conclusion}

{
    \small
    \bibliographystyle{ieeenat_fullname}
    \bibliography{main}
}

\clearpage
\newpage

\section*{Contents}
  \vspace{12pt}
The following items are included in the supplementary material:
  \vspace{12pt}
\begin{itemize}
  \item Quantitative Evaluation of both CIRR and Fashion-IQ in Section~\ref{sec:qua}.
  \vspace{10pt}
  \item GPT-4 Prompt Example for Generating QA Pairs in Section~\ref{sec:pro}.
  \vspace{10pt}
  \item Failure Cases Generated from GPT-4 in Section~\ref{sec:failure}.
  
    \vspace{10pt}
\end{itemize}
%

\section{Quantitative Evaluation}\label{sec:qua}
\begin{figure*}[t]
	\begin{center}
		\includegraphics[width=\linewidth]{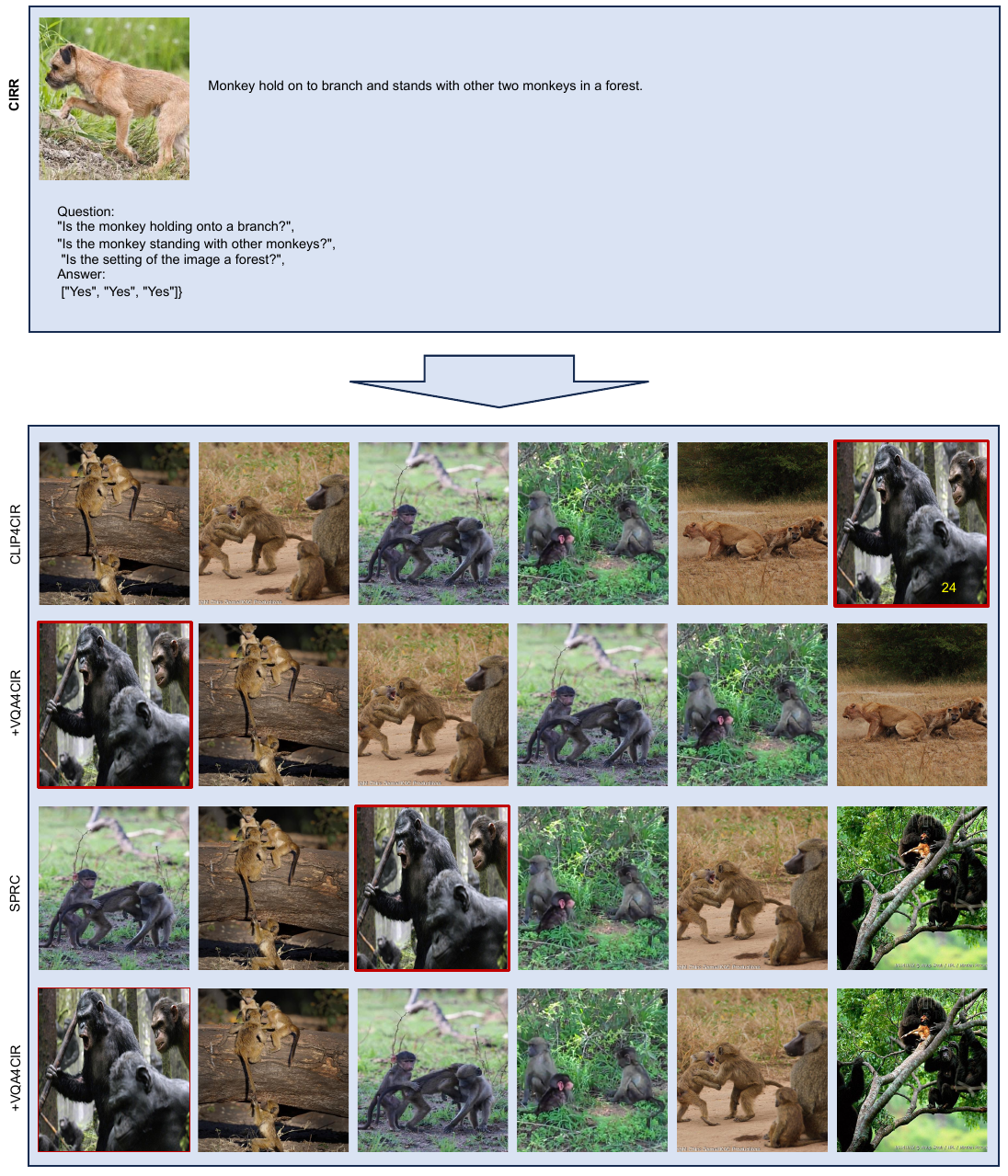}
	\end{center}
	\vspace{-13pt}
	\captionsetup{font=small}
	\caption{\textbf{Qualitative} analysis of case \textbf{(a)} of CLIP4CIR~\cite{baldrati2023composed}, SPRC~\cite{bai2023sentence} and our VQA4CIR along with the generated QA pairs on \textbf{CIRR} dataset, where the target images are highlighted with \texttt{{\color{firebrick}red}} outline. We note that image dimensions are adjusted for aesthetic layouts due to space constraints.}
	\label{fig:qua}
\end{figure*}

\begin{figure*}[t]
	\begin{center}
		\includegraphics[width=\linewidth]{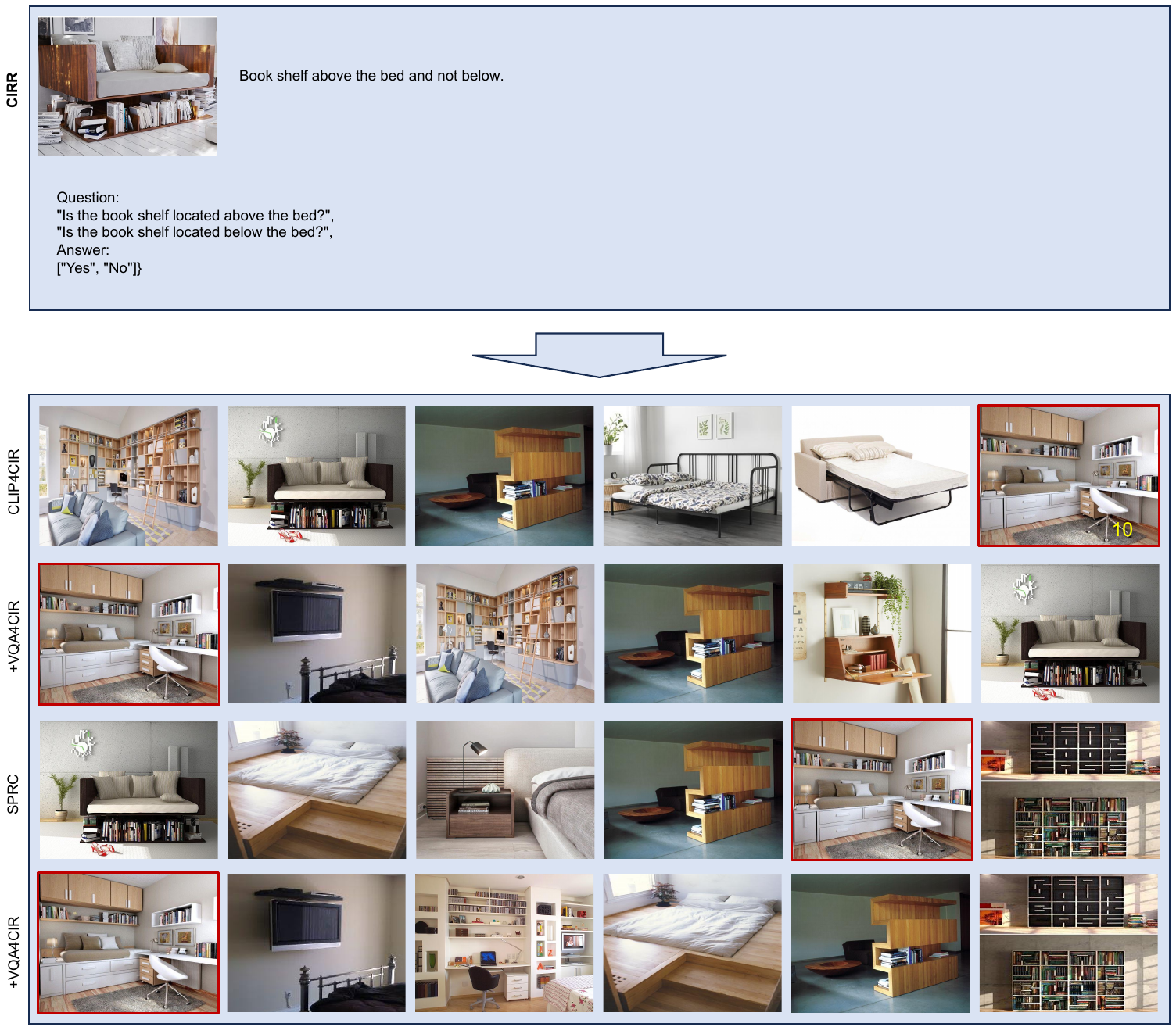}
	\end{center}
	\vspace{-13pt}
	\captionsetup{font=small}
	\caption{\textbf{Qualitative} analysis of case \textbf{(b)} CLIP4CIR~\cite{baldrati2023composed}, SPRC~\cite{bai2023sentence} and our VQA4CIR along with the generated QA pairs on \textbf{CIRR} dataset, where the target images are highlighted with \texttt{{\color{firebrick}red}} outline. We note that image dimensions are adjusted for aesthetic layouts due to space constraints.}
	\label{fig:qua1}
\end{figure*}

\begin{figure*}[t]
	\begin{center}
		\includegraphics[width=\linewidth]{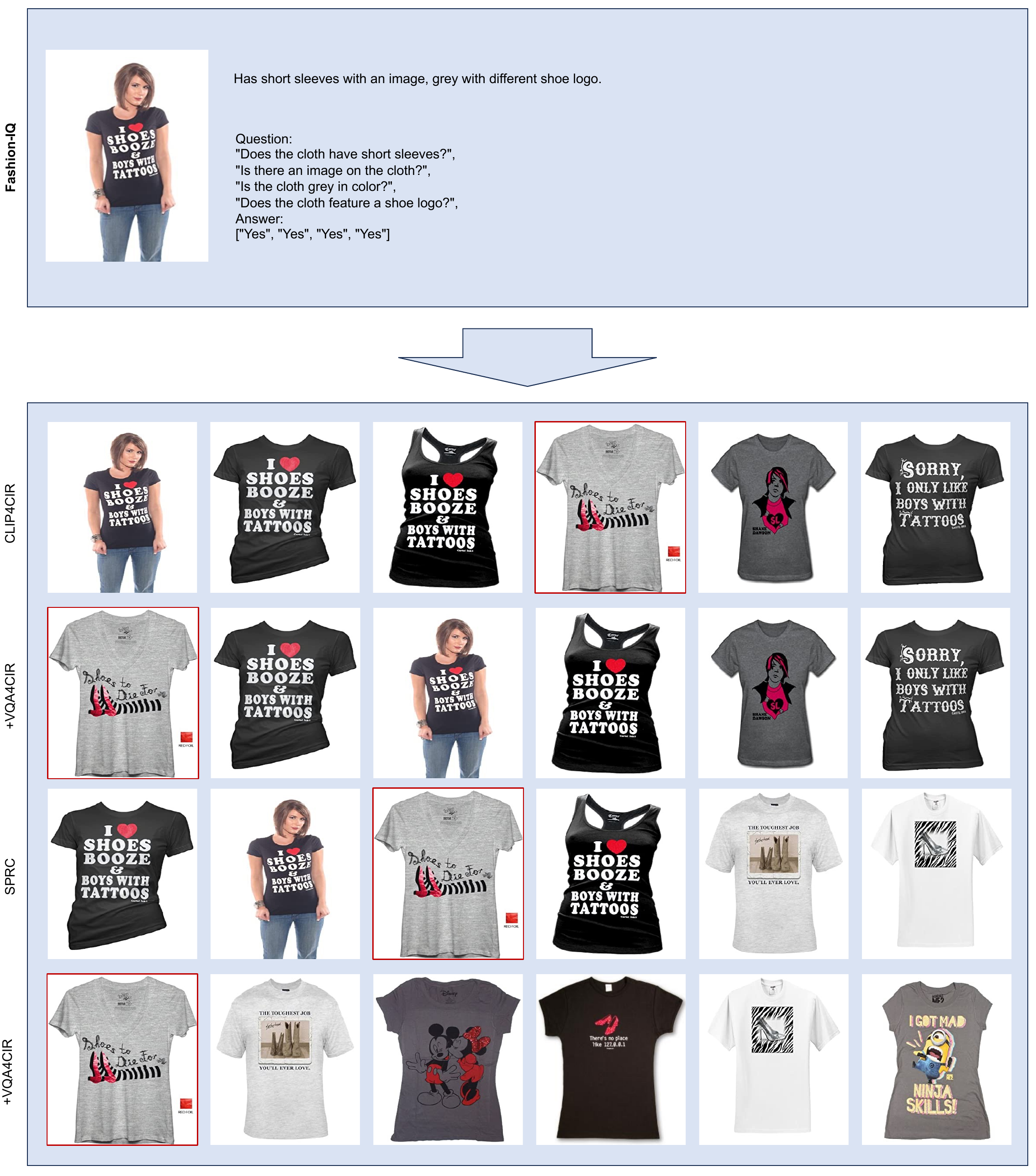}
	\end{center}
	\vspace{-13pt}
	\captionsetup{font=small}
	\caption{\textbf{Qualitative} analysis of case \textbf{(c)} CLIP4CIR~\cite{baldrati2023composed}, SPRC~\cite{bai2023sentence} and our VQA4CIR along with the generated QA pairs on \textbf{Fashion-IQ} dataset, where the target images are highlighted with \texttt{{\color{firebrick}red}} outline. We note that image dimensions are adjusted for aesthetic layouts due to space constraints.}
	\label{fig:qua2}
\end{figure*}

\begin{figure*}[t]
	\begin{center}
		\includegraphics[width=\linewidth]{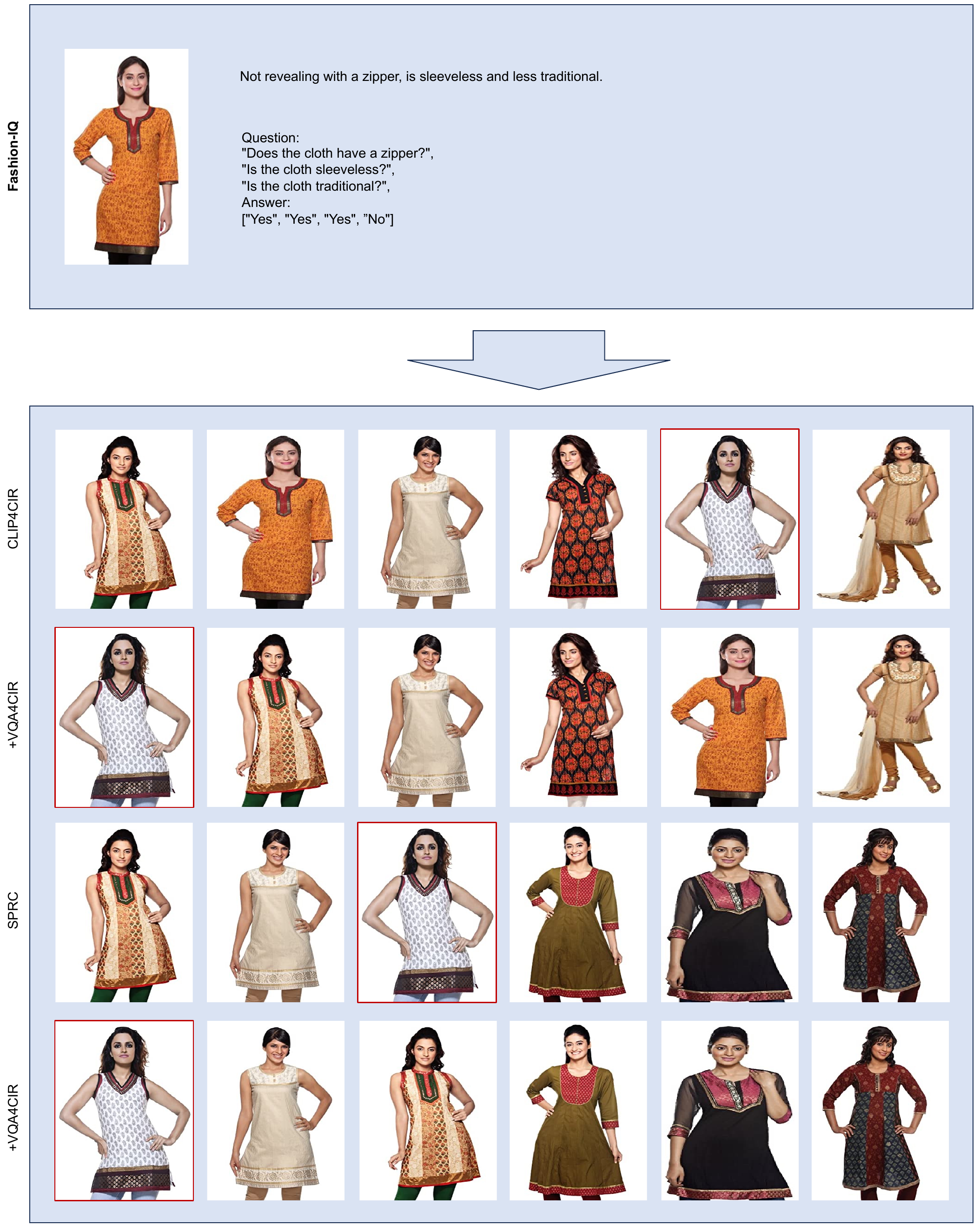}
	\end{center}
	\vspace{-13pt}
	\captionsetup{font=small}
	\caption{\textbf{Qualitative} analysis of case \textbf{(d)} CLIP4CIR~\cite{baldrati2023composed}, SPRC~\cite{bai2023sentence} and our VQA4CIR along with the generated QA pairs on \textbf{Fashion-IQ} dataset, where the target images are highlighted with \texttt{{\color{firebrick}red}} outline. We note that image dimensions are adjusted for aesthetic layouts due to space constraints.}
	\label{fig:qua3}
\end{figure*}

To quantitatively evaluate the effectiveness of our method, we visualized the re-ranking results of VQA4CIR with respect to the Recall@Subset1 metric for two baseline methods across two datasets, as shown in Figs.~\ref{fig:qua}, ~\ref{fig:qua1}, ~\ref{fig:qua2}, and ~\ref{fig:qua3}.
Target images are highlighted with \texttt{{\color{firebrick}red}} boxes, where a higher ranking of the target image indicates better retrieval performance and vice versa.
The visualizations show that, by incorporating VQA for excluding the candidate images being inconsistent with relative captions, there is a notable improvement in the retrieval of target images.
For instance, in Fig.~\ref{fig:qua} of case \textbf{(a)} from the {CIRR} dataset, VQA4CIR deconstructs the relative caption into three questions: ``\texttt{Is the monkey holding onto a branch}?'', ``\texttt{Is the monkey standing with other monkeys}?'', and ``\texttt{Is the setting of the image a forest}?''.
These questions grasp the key issues of the relative caption, \ie, holding onto a branch, standing with other monkeys, and being in a forest, using VQA to assist the model effectively captures the key points of relative caption.
From the retrieval results of CLIP4CIR~\cite{baldrati2023composed} and SPRC~\cite{bai2023sentence}, one can see that the top-ranked candidate images do not fully satisfy the descriptions involved in the three questions.
Notably, CLIP4CIR ranked the target image in the $24$-th position. Interestingly, our method ranks the target image first, regardless of whether it is based on CLIP4CIR~\cite{baldrati2023composed} or SPRC~\cite{bai2023sentence}.

Moreover, we observed a similar effect in cases Fig.~\ref{fig:qua2} of \textbf{(c)} and Fig.~\ref{fig:qua3} of \textbf{(d)} on the {Fashion-IQ} dataset.
For example, in case \textbf{(c)}, the relative caption mentioned ``\texttt{Has short sleeves with an image, grey with different shoe logo}?'', which contains three characteristics, \ie, short sleeves, grey color, and shoe logo.
With this perspective, our VQA4CIR deconstructs those characteristics into four distinct sub-questions, \ie, ``\texttt{Does the cloth have short sleeves}?'', ``\texttt{Is there an image on the cloth}?'', ``\texttt{Is the cloth grey in color}?'', and ``\texttt{Does the cloth feature a shoe logo}?''
For each candidate image, we  evaluate it against the four specified questions. Images that fail to conform are relegated in sequence, while those that comply with all the criteria are identified as prospective target images and are accordingly moved up in priority.
Given this, VQA4CIR obtains the final ranking by retrieving candidate images and answering these questions one by one.
Compared to CLIP4CIR~\cite{baldrati2023composed} and SPRC~\cite{bai2023sentence}, VQA4CIR can accurately rank the candidate images that satisfy all the characteristics involved in the relative caption first, while CLIP4CIR~\cite{baldrati2023composed} and SPRC~\cite{bai2023sentence} rank the target image at the $4$ and $3$ positions, respectively.
The results demonstrate that using the VQA mechanism to address the challenges of the CIR task is effective.

\section{GPT-4 Prompt Example for Generating QA Pairs}\label{sec:pro}
\begin{figure*}
    \centering
    \includegraphics[width=\linewidth]{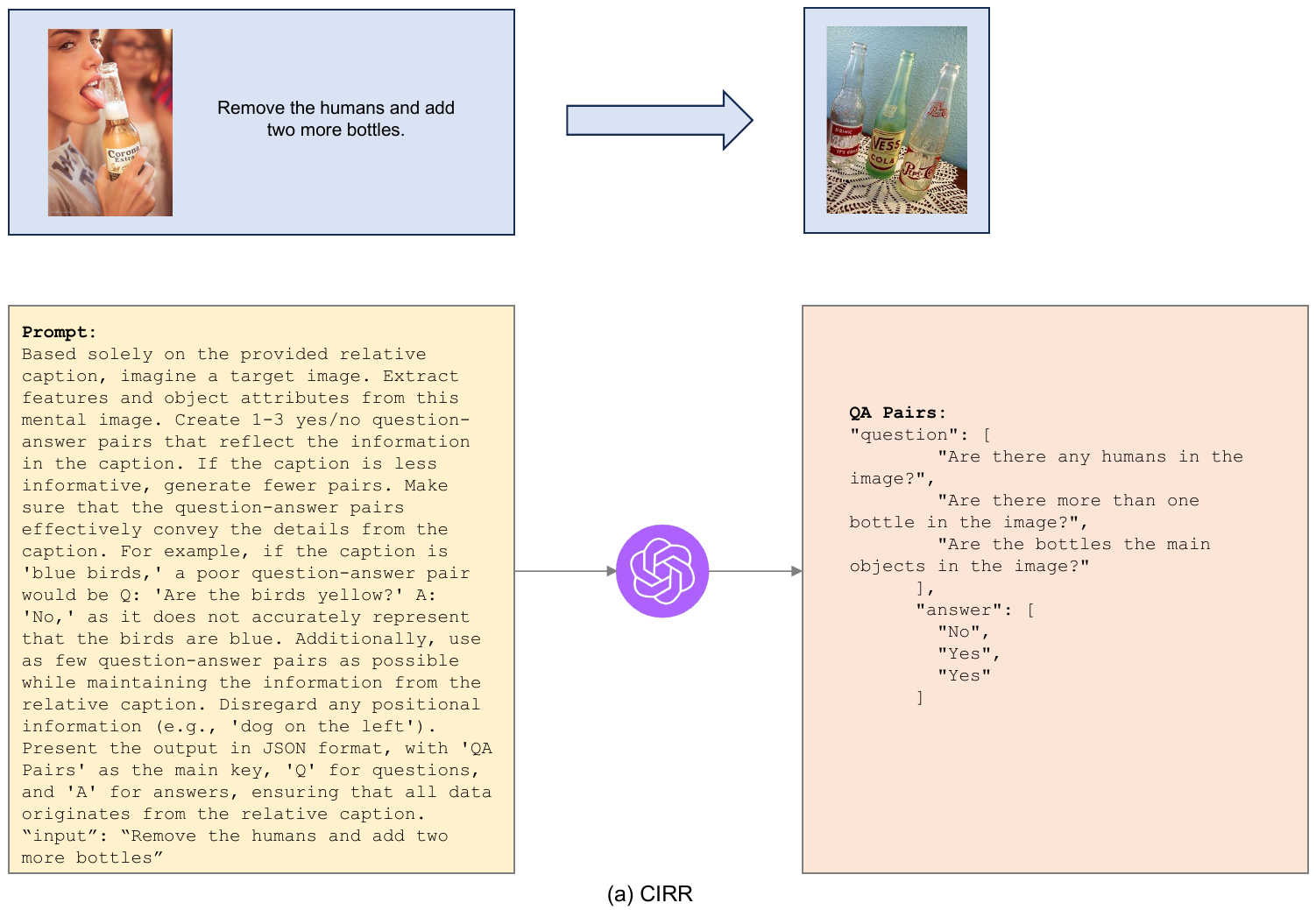}
    \caption{Example of a GPT-4 prompt for generating QA pairs on the CIRR dataset, along with the corresponding reference image, relative caption, and target image are provided.}
    \label{fig:prom1}
\end{figure*}

\begin{figure*}
    \centering
    \includegraphics[width=\linewidth]{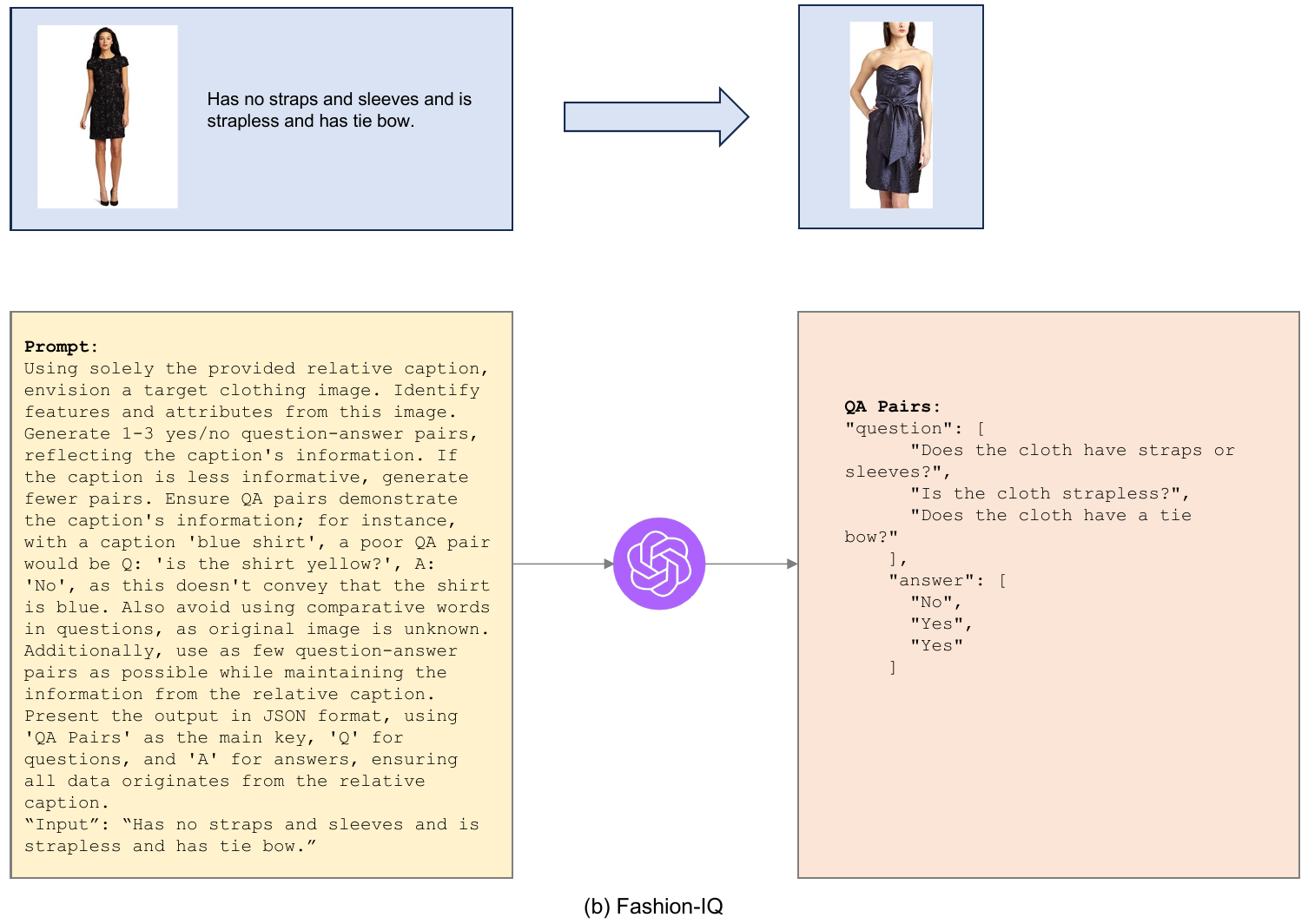}
    \caption{Example of a GPT-4 prompt for generating QA pairs on the Fashion-IQ dataset, along with the corresponding reference image, relative caption, and target image are provided.}
    \label{fig:prom2}
\end{figure*}

Here we display several examples of GPT-4 prompts for generating instruction data, \ie, QA pairs, in Fig.~\ref{fig:prom1} and \ref{fig:prom2}.
For CIRR, as shown in Figs.~\ref{fig:prom1}, our goal is to ask GPT-4 to imagine a target image based solely on the provided relative caption, and create $1\sim3$ yes/no questions that respond to the relative caption. 
To ensure the questions to be more informative, we give GPT-4 an example, \eg, if the caption is `\texttt{blue birds},' a poor question-answer pair would be \textbf{Q}: `\texttt{Are the birds yellow?}' \textbf{A}: `\texttt{No},' as we cannot deduce \emph{the birds are blue} from this QA pair. 
Then, we specifically request GPT-4 to describe the information in the relative caption using the minimum number of QA pairs to prevent redundancy. Finally, for the relative caption `Remove the humans and add two more bottles', GPT-4 generated the three QA pairs illustrated in Fig.~\ref{fig:prom1}. Based on the reference and target images in Fig.~\ref{fig:prom1}, we can see that these QA pairs effectively reflect the key characteristics in relative caption.

Similarly, as shown in Fig.~\ref{fig:prom2}, to make GPT-4 generate QA pairs solely based on the relative caption provided by Fashion-IQ, we gave a counter-example, \eg, with a caption `\texttt{blue shirt}', a poor QA pair would be Q: `\texttt{Is the shirt yellow}?', A: `\texttt{No}', as one cannot deduce \emph{the shirt is blue} from the QA pair. Additionally, we explicitly request GPT-4 to avoid generating this kind of question. Finally, all question-and-answer pairs are presented in JSON format, with `QA Pairs' as the main key, `\textbf{Q}' for the question, and `\textbf{A}' for the answer, ensuring all data is derived from the relative caption.

\section{Failure Cases Generated from GPT-4} \label{sec:failure}
\begin{figure*}
    \centering
    \includegraphics[width=\linewidth]{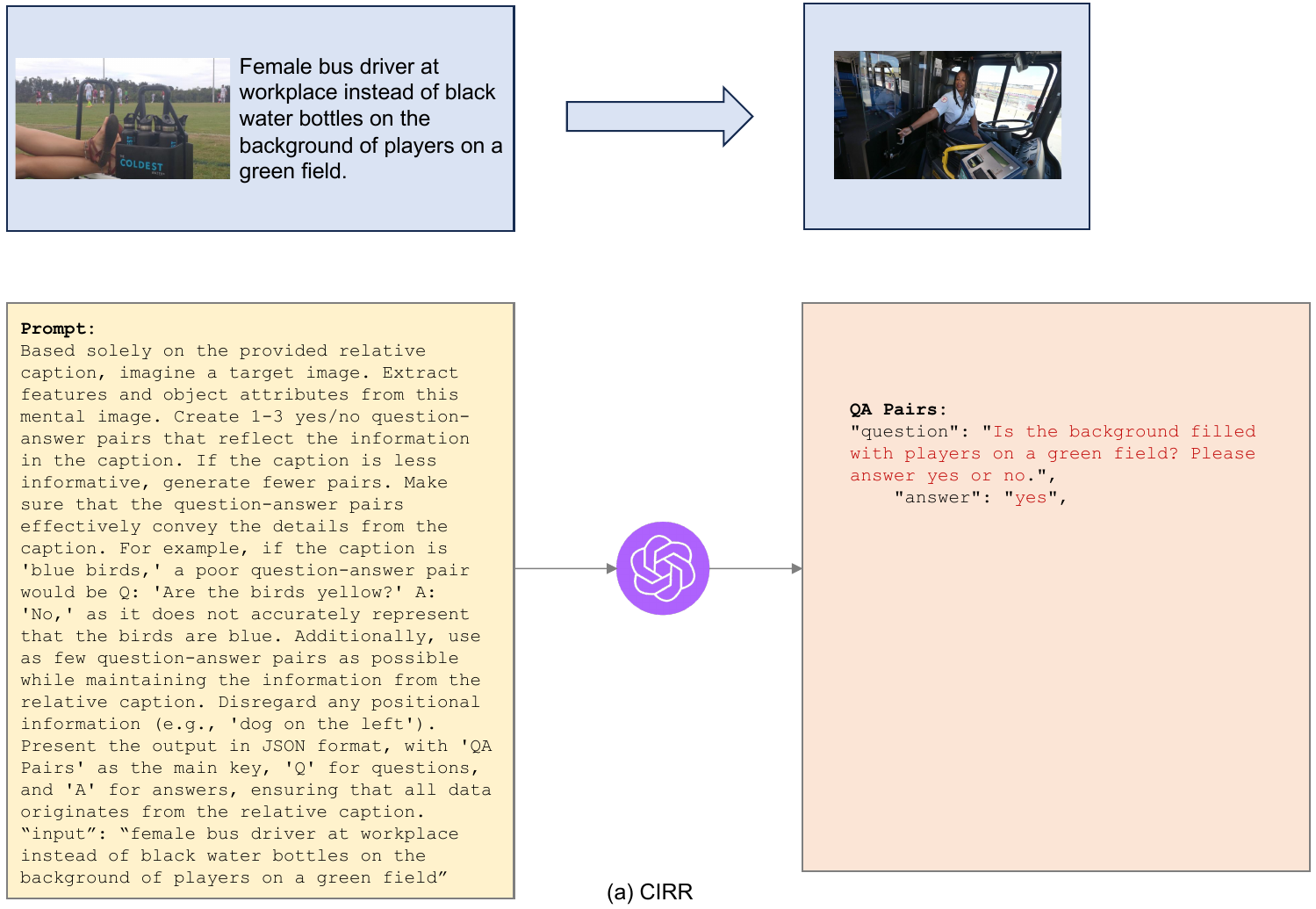}
    \caption{Failure examples generated from GPT-4 on the CIRR dataset, along with the corresponding reference image, relative caption, and target image are provided.}
    \label{fig:failure}
\end{figure*}

\begin{figure*}
    \centering
    \includegraphics[width=\linewidth]{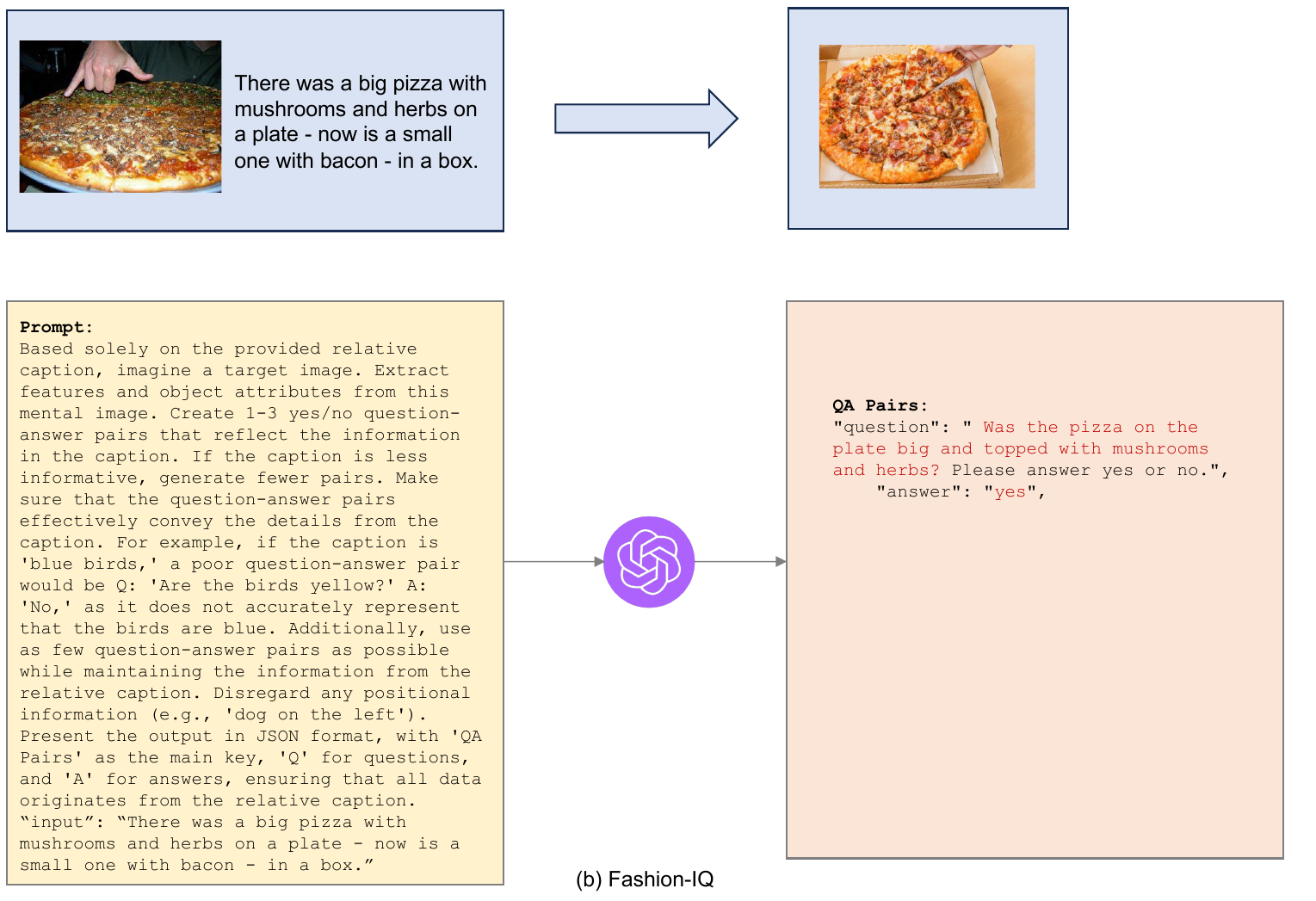}
    \caption{Failure examples generated from GPT-4 on the Fashion-IQ dataset, along with the corresponding reference image, relative caption, and target image are provided.}
    \label{fig:failure2}
\end{figure*}

While the above prompt can generate a large number of good-quality QA pairs on the CIRR and Fashion-IQ datasets, there are still some instances of failure that require manual refinement. As shown in Fig.~\ref{fig:failure}, the relative caption requests to search an image where `\texttt{female bus driver at workplace instead of black water bottles on the background of players on a green field}'. However, the question generated by GPT-4 is: `\texttt{Is the background filled with players on a green field}?'  In this case, the target image would not possibly have a green field.  Therefore, this case requires manual correction. Similarly, in Fig.~\ref{fig:failure2}, we observe that the content described in the questions generated by GPT-4 is all relevant to the content in the reference image and does not directly relate to the target image.

\end{document}

%% file: preamble.tex
\usepackage[dvipsnames]{xcolor}


%% file: sec/0_Abstract.tex
\begin{abstract}
Albeit progress has been made in Composed Image Retrieval (CIR), we empirically find that a certain percentage of failure retrieval results are not consistent with their relative captions.
To address this issue, this work provides a Visual Question Answering (VQA) perspective to boost the performance of CIR.
The resulting VQA4CIR is a post-processing approach and can be directly plugged into existing CIR methods.
%
%
Given the top-$C$ retrieved images by a CIR method, VQA4CIR aims to decrease the adverse effect of the failure retrieval results being inconsistent with the relative caption.
To find the retrieved images inconsistent with the relative caption, we resort to the "QA generation $\rightarrow$ VQA" self-verification pipeline.
For QA generation, we suggest fine-tuning LLM (\eg, LLaMA) to generate several pairs of questions and answers from each relative caption. 
We then fine-tune LVLM (\eg, LLaVA) to obtain the VQA model.
By feeding the retrieved image and question to the VQA model, one can find the images inconsistent with relative caption when the answer by VQA is inconsistent with the answer in the QA pair.
Consequently, the CIR performance can be boosted by modifying the ranks of inconsistently retrieved images.
Experimental results show that our proposed method outperforms state-of-the-art CIR methods on the CIRR and Fashion-IQ datasets.
%
\end{abstract}

%% file: sec/1_Intro.tex
\section{Introduction}
\label{sec:intro}

Composite Image Retrieval (CIR)~\cite{liu2021image,vo2019composing,baldrati2022effective} is a challenging retrieval task, where a reference image together with a relative caption are combined to retrieve the desired target image.
%
%
Benefiting from its dual-modal query nature, CIR can offer a nuanced depiction of the desired image, allowing for interactive refinements by tweaking the reference image and description.
Such capabilities make CIR especially suited for applications like e-commerce and digital search~\cite{feng2020learning}.

\begin{figure}[t]
	\begin{center}
		\includegraphics[width=\linewidth]{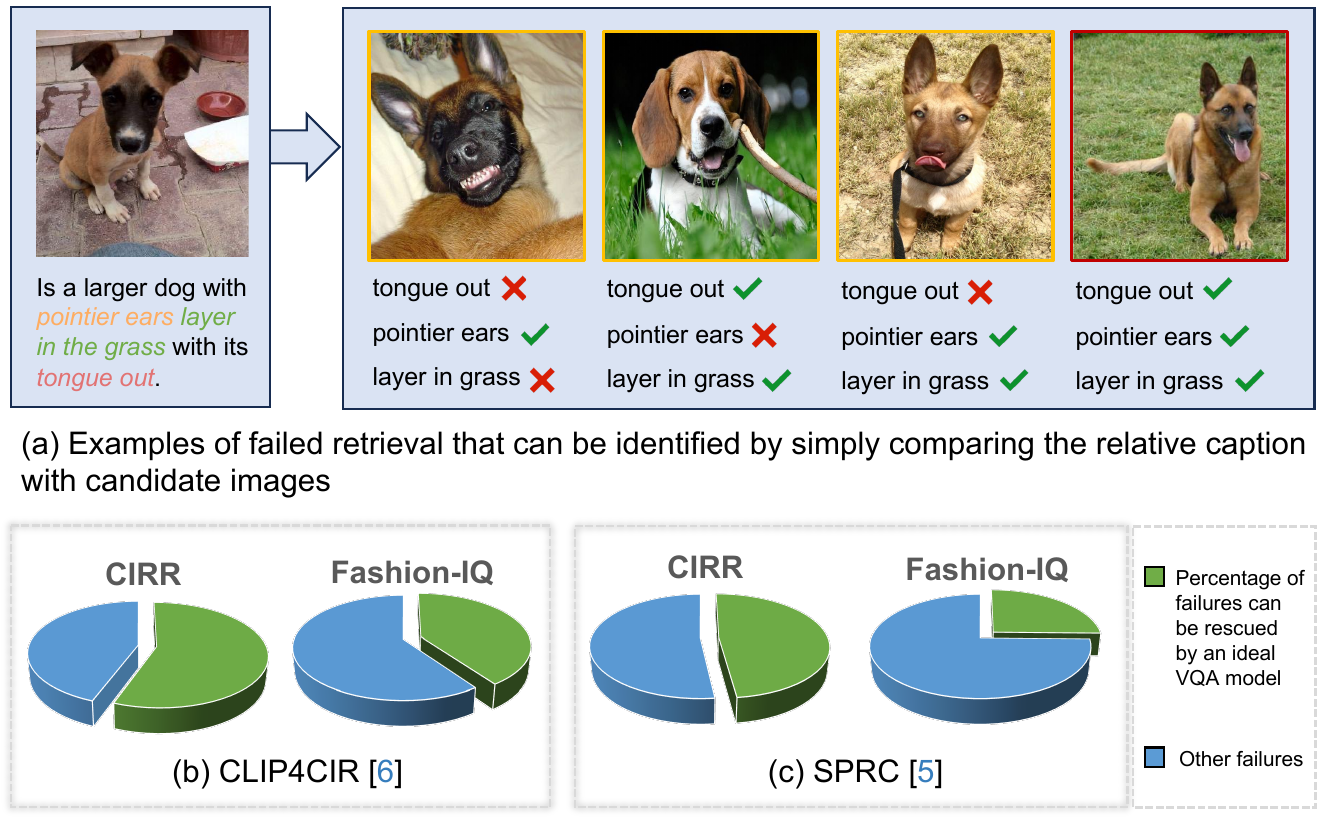}
	\end{center}
	\vspace{-15pt}
	\captionsetup{font=small}
	\caption{\small \textbf{Failure cases analysis}. \textbf{(a)} Examples of failure retrieval results of SPRC~\cite{bai2023sentence} on CIRR dataset, which one can see that they can be identified by comparing with relative caption, \textbf{(b)} and \textbf{(c)} are the statistical analysis of the percentage of failures from CLIP4CIR~\cite{baldrati2022conditioned} and SPRC~\cite{bai2023sentence} on the CIRR and Fashion-IQ datasets, where one can see that a \textit{certain percentage} of failure cases can be ascribed to the \textit{inconsistency} between retrieved images and relative captions.}
	\vspace{-19pt}
	\label{fig:intro}
\end{figure}

%
%
%
%
%
%

In the recent few years, considerable progress has been made in CIR.
For example, many late fusion methods~\cite{anwaar2021compositional,chen2020image,dodds2020modality,liu2021image,vo2019composing} have been developed to integrate the information from reference image and relative caption.
Pseudo-word embedding~\cite{gal2022image,baldrati2022effective} is shown to convert reference image into description embedding for combining with the relative caption to retrieve the target image.
Taking both reference images and relative captions into account, sentence-level prompts have also been proposed to enhance CIR~\cite{bai2023sentence}.

However, for most existing CIR methods, we empirically find that a certain percentage of retrieval results are not consistent with their relative captions.
For example, in Fig.~\ref{fig:intro} (a), given the relative caption ``\texttt{Is a larger dog with pointier ears laying in the grass with its tongue out}'', the rank-1 image retrieved by SPRC~\cite{bai2023sentence} does not satisfy the attributes \texttt{{\textcolor[RGB]{118,193,118}{\texttt{{lying in the grass}}}}}, and \texttt{{\textcolor[RGB]{225,115,115}{\texttt{{tongue out}}}}}.
In Fig.~\ref{fig:intro} (b) and (c), we summarize the percentage of causes of failures from randomly selected $150$ failure cases of the state-of-the-art methods CLIP4CIR~\cite{baldrati2022conditioned} and SPRC~\cite{bai2023sentence} from the validation set on CIRR and Fashion-IQ datasets.
Among these, $55.8$\% and $48.2$\% of failures on CIRR, $38.3$\% and $25.3$\% of the failures on Fashion-IQ, can be attributed to the inconsistency between the retrieved images and relative captions.
Motivated by the above result, this paper aims to present a post-processing method to find the retrieved images that are inconsistent with the relative captions, and to modify their ranks for boosting CIR performance.

\begin{figure}[t]
	\vspace{-4pt}
	\begin{center}
		\includegraphics[width=\linewidth]{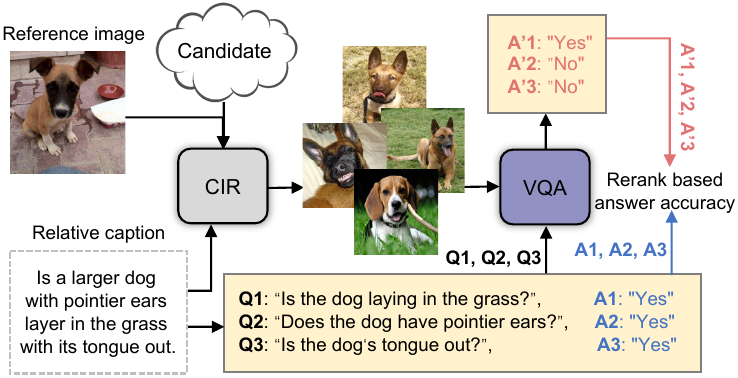}
	\end{center}
	\vspace{-15pt}
	\captionsetup{font=small}
	\caption{\small \textbf{Illustration} of the \textit{\textbf{main idea}} of VQA4CIR, which converts the relative captions into multiple QA pairs, and uses the VQA model to respond to each candidate image. Finally, \textit{{reranking}} is conducted on the candidate images by comparing the answers of the VQA model and QA pairs.}
	\vspace{-19pt}
	\label{fig:intro2}
\end{figure}

To find the retrieved images being inconsistent with relative captions, we resort to the "QA generation $\rightarrow$ VQA" self-verification pipeline, resulting in our VQA4CIR method (see Fig.~\ref{fig:intro2}).
In QA generation, we aim to generate question $\textbf{Q}_i$ and answer $\textbf{A}_i$ pairs from relative caption.
Motivated by the unprecedented success of large language models (LLMs), we adopt the open-sourced LLaMA~\cite{touvron2023llama} and fine-tune it to fulfill our requirements.
To construct the instruction dataset, we use GPT-4~\cite{OpenAI2} to generate QA pairs, and modify parts of low-quality QA pairs in a handcrafted manner.
Using the instruction dataset, we use the pre-defined instruction prompt, and adopt LoRA~\cite{hu2021lora} for fine-tuning LLaMA~\cite{touvron2023llama}.
In VQA, we feed each question $\textbf{Q}_i$ and the retrieved image into the VQA model to generate an answer ${\textbf{A}}'_i$.
For training the VQA model, we fine-tune large vision-language models (\eg, LLaVA~\cite{liu2023visual}) using the instruction data.
When all ${\textbf{A}}'_i$s are equal to the corresponding ${\textbf{A}}_i$s, the retrieved image will be regarded to be consistent with relative caption.
By modifying the ranks of the inconsistent retrieved images, we rerank the retrieved images to boost CIR performance.

Our proposed VQA4CIR is a post-processing approach and can be directly plugged into existing CIR methods for better CIR performance.
Using the top-4 retrieved results by SPRC~\cite{bai2023sentence} in Fig.~\ref{fig:intro} (a) as an example, from the relative caption, one can use LLM to generate three QA pairs in Fig.~\ref{fig:intro2}.
Obviously, for each of the three top-rank retrieved images, at least one of the answers is not consistent with the answer in the QA pairs.
For all questions, the answers to the target image are consistent with the answers in the QA pairs.
Thus, we can modify the ranks of the first three retrieved images to be later compared with the target image, thereby improving the CIR performance.
Extensive experiments are conducted on the CIRR and Fashion-IQ datasets.
The results show that our VQA4CIR can be incorporated with different CIR methods and outperforms the state-of-the-art CIR methods.

To sum up, the contributions of this work are three-fold:
\begin{itemize}[leftmargin=*]
	\setlength{\itemsep}{0pt}
	\setlength{\parsep}{-2pt}
	\setlength{\parskip}{-0pt}
	\setlength{\leftmargin}{-15pt}
\item 
By analyzing the failure CIR retrieved results, we suggest a \textit{VQA perspective} for boosting the performance of existing CIR approaches, resulting in our VQA4CIR method. 
%
   
\item 
Following the "\textit{QA generation $\rightarrow$ VQA}" \textit{self-verification pipeline}, we fine-tune LLM to generate QA pairs from the relative caption, and fine-tune LVLM to answer the questions. By finding and modifying the ranks of the inconsistent retrieved images, we rerank the retrieved images to attain better CIR performance.
%


\item 
Experimental results show that our VQA4CIR  outperforms the state-of-the-art CIR methods and can be directly plugged into existing CIR methods.
\vspace*{-7pt}
\end{itemize}

%% file: sec/2_RelatedWork.tex
\section{Related Work}
\label{sec:relatedwork}

\noindent\textbf{Composed Image Retrieval.}
Early CIR techniques primarily employed a late fusion strategy to combine features from a reference image with its relative caption. Then, the merged features were compared with all candidate image features to retrieve the most matched image among all candidates in an extensive image corpus ~\cite{anwaar2021compositional,chen2020image,dodds2020modality,liu2021image,vo2019composing}.
Subsequent developments introduced various feature fusion methods~\cite{vo2019composing} and attention mechanisms~\cite{chen2020image,dodds2020modality}, and exhibited impressive performance in CIR tasks.
Recently, many CIR techniques began to leverage the abilities of pre-trained models to enhance image and text features~\cite{baldrati2022effective,ventura2023covr,baldrati2022conditioned,liu2023candidate,gu2023compodiff}.
For example, FashionVLP~\cite{goenka2022fashionvlp} utilized the pre-trained BERT~\cite{devlin2018bert} to integrate triplet data, \ie, image, text, and tag features. 
Cola~\cite{ray2023cola} leveraged pre-trained CLIP to retrieve the relationship between objects and attributes using two kinds of query approaches, \ie, single- and multi-object. 
Liu \textit{et al.} employed a two-stage approach on top of the pre-trained model to retrieve target images, \ie, first rapidly screened candidate objects and then re-ordered these candidates~\citep{liu2023bi}.
Another line of techniques adopted the 'text inversion' approach to transform a reference image into pseudo-word embeddings, which are then combined with their relative captions, facilitating text-to-image retrieval.
Liu \textit{et al}. constructed texts with semantics opposite to the original and embedded learnable tokens within, allowing image retrieval in two distinct queries~\cite{liu2023candidate}.
Bai \textit{et al}. employed both the reference image and relative caption to derive sentence-level prompts, aiming to enrich the caption by offering a proper description of pertinent elements within the reference image~\cite{bai2023sentence}.
However, existing CIR methods are limited in retrieving the target images, and a certain percentage of failure retrieval results even are not consistent with their relative captions (see Fig.~\ref{fig:intro}).
%
%
Thus, we suggest using VQA4CIR to find these inconsistent images and rerank the retrieved images.
%
%
While Liu \textit{et al}. also aims to mitigate errors from the first stage through a re-ranking mechanism~\cite{liu2023candidate}, it merely concatenates two traditional CIR methods without addressing the underlying issue of CIR.


\begin{figure*}[!t]
    \vspace{-11pt}
	\begin{center}
		\includegraphics[width=\linewidth]{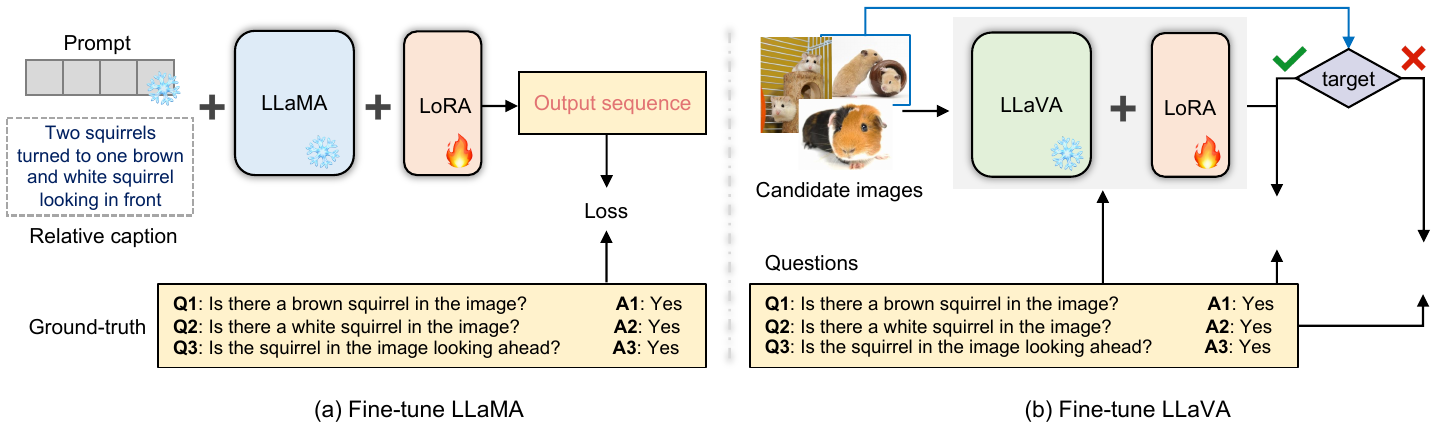}
  \put(-96,67){\tiny$-\log\!\left(p_1\!\cdot \!p_2\!\cdot\!\ldots\!p_K\!\right)$}
  \put(-50,53){\tiny$-\log\!\left(\!1\!-\!\left(p_1\!\cdot\!p_2\!\cdot\!\ldots\!p_K\!\right)\right)$}
	\end{center}
    \vspace{-16pt}
	\captionsetup{font=small}
	\caption{\small\textbf{Overview} of the \textit{training stage} of our \textbf{VQA4CIR}. \textbf{(a)} Fine-tuning the LLaMA~\cite{touvron2023llama} with the instruction data, where the prompt and backbone are frozen while the LoRA is learnable. \textbf{(b)} Fine-tuning the LLaVA~\cite{liu2023visual} with the training data, where the backbone is frozen and only the LoRA is learnable. 
 }
	\vspace{-18pt}
	\label{fig:3}
\end{figure*}

\noindent\textbf{LLMs and LVLMs.}
In the recent few years, LLMs have demonstrated remarkable capabilities in language generation, contextual learning, world knowledge, and reasoning.
The GPT familes, \eg, GPT-3~\cite{brown2020language}, ChatGPT~\cite{OpenAI1}, GPT-4~\cite{OpenAI2}, and InstructGPT~\cite{ouyang2022training}, stands as the most significant achievements in LLMs.
Open-source models like OPT~\cite{zhang2022opt}, LLaMA~\cite{touvron2023llama}, MOSS~\cite{sun2023moss}, Alpaca~\cite{taori2023stanford}, Vicuna~\cite{chiang2023vicuna}, and GLM~\cite{du2022glm} served as valuable resources that allowed fine-tuning for specific domains or tasks.
%
Most recently, a plethora of research has focused on extending LLMs into LVLMs, \eg, Minigpt-4~\cite{zhu2023minigpt}, LLaVA~\cite{liu2023visual}, instructBLIP~\cite{dai2023instructblip}.
Typically, LVLMs comprise a visual encoder, a language encoder (\ie, LLM), and a cross-modal alignment network.
%
%
Many efficient vision-text interactions~\cite{li2023blip}, efficient training methods~\cite{gao2023llama,zhang2023transfer}, and instruction tuning~\cite{liu2023visual,zhu2023minigpt,ye2023mplug,dai2023instructblip} methods have been proposed.
%
%
%
%
LVLMs are adept at intricate reasoning surrounding these objects, culminating in enhanced outcomes in diverse multimodal challenges by Visual Question Answering (VQA).
As such, we resort to the powerful performance of LLM and LVLM to conduct complex reasoning on relative captions and retrieved images through QA generation and VQA, thereby offering a new perspective for improving VQA performance.

\noindent\textbf{Downstreaming Applications of LLMs and LVLMs.} 
LLMs and LVLMs have demonstrated exceptional and innovative performance in various downstream few-shot vision learning scenarios.
For example, in the realm of object detection, leveraging prompt embeddings to fine-tune the LVLMs via prompt learners can lead to cutting-edge results~\cite{gu2023anomalygpt}.
In image segmentation, LVLMs not only enhance the performance of open vocabulary but also allow the segmentation model to inherit and utilize the language generation capabilities of LLM~\cite{lai2023lisa,bangalath2022bridging,liang2023open}.
For video understanding, LVLMs can help boost the capability of understanding and generating human-like conversations about videos~\cite{li2023videochat,maaz2023video}.
For 3D representations, LVLMs features can be used to encode semantics in 3D representations~\cite{gu2023conceptgraphs}.
Inspired by the outstanding performance of LLMs and LVLMs in various domains, this work explores the potential of leveraging LLM and LVLM to enhance the CIR performance.

%% file: sec/3_Methodology.tex
\section{Methodology}\label{sec:method}

\noindent\textbf{Overview.}~
In CIR, the multimodal composite query $\{\boldsymbol I_r, \boldsymbol t\}$ involves a reference image $\boldsymbol I_r$   and a  relative caption $\boldsymbol t$.
The essence of CIR lies in retrieving the target image from an extensive image corpus $\mathcal{D}$, relying on the content of both the reference image and relative caption.
Given that the target image needs to capture the changes in objects and attributes described in relative caption while preserving visual resemblances to the reference image, this positions CIR as more challenging than conventional text-to-image retrieval.

In this work, we suggest a new perspective on CIR through the lens of VQA, resulting in our VQA4CIR.
VQA4CIR can be incorporated with any existing CIR methods.
First, by finetuning LLM, we obtain a QA generation model to generate QA pairs $\{(\textbf{Q}_1, \textbf{A}_1), ..., (\textbf{Q}_K, \textbf{A}_K)\}$ from the relative caption $\boldsymbol t$. 
Using any CIR method, we can get its top-$C$ rank candidate images $\{\boldsymbol{I}_1, ...,  \boldsymbol{I}_c, ..., \boldsymbol{I}_C\}$. 
Then, by finetuning LVLM, we obtain the VQA model.
For each retrieved image $\boldsymbol{I}_c$ and question $\textbf{Q}_k$, the VQA model takes $(\boldsymbol{I}_c, \textbf{Q}_k)$ as the input to generate the answer $\textbf{A}'_k$. 
When $\textbf{A}'_k$ is not equal to $\textbf{A}_k$, we can treat $\boldsymbol{I}_c$ to be inconsistent with relative caption and modify its rank.
In this way, the adverse effect of inconsistent retrieved images can be suppressed and better CIR performance can be attained.
In the following, we will introduce the QA generation, VQA, and the inference process in detail.

%
%
%
%
%

\subsection{Generating QAs from Relative Caption}\label{sec:llama}
%
LLM is powerful in many natural language processing tasks but should be finetuned to match the requirements of VQA4CIR in generating QA pairs from relative captions.
To this end, we construct an instruction dataset and finetune LLaMA as follows.


\noindent\textbf{Construction of Instruction Dataset.}~
To train LLaMA~\cite{touvron2023llama} for generating QA pairs from relative captions, we require a substantial number of question-and-answer pairs $\mathbf{x}_{\texttt{QA}s}$ as training data.
%
%
Motivated by the great success of GPT models in text generation~\cite{gilardi2023chatgpt}, we employ GPT-4 to ease the cost of labor-intensive manual annotations of QA pairs.
%
%
In general, the generated QA pairs should cover all the contents of the relative caption and should have a small number.
We also recommend the answer to be `\texttt{yes}' or `\texttt{no}' for clarity. 
To fulfill these requirements, we formulated a set of prompts for GPT-4 (refer to the \textit{Suppl.} for further details).
%
%
%
%
%
%
%
Nonetheless, a small percentage of QA pairs generated by GPT-4 are of poor quality.
And we further conducted a manual review to refine and remove them.
The resulting instruction dataset is presented in \textbf{\texttt{JSON}} format, with `\texttt{QA Pairs}' as the main key, `Q' for questions, and `A' for answers, ensuring that all data originates from the relative caption.
%
%
In our experiments, we selected $5,000$ and $3,000$ samples to construct the instruction data for the {CIRR} and {Fashion-IQ} datasets, respectively.

\begin{figure}[!t]
	\begin{center}
		\includegraphics[width=\linewidth]{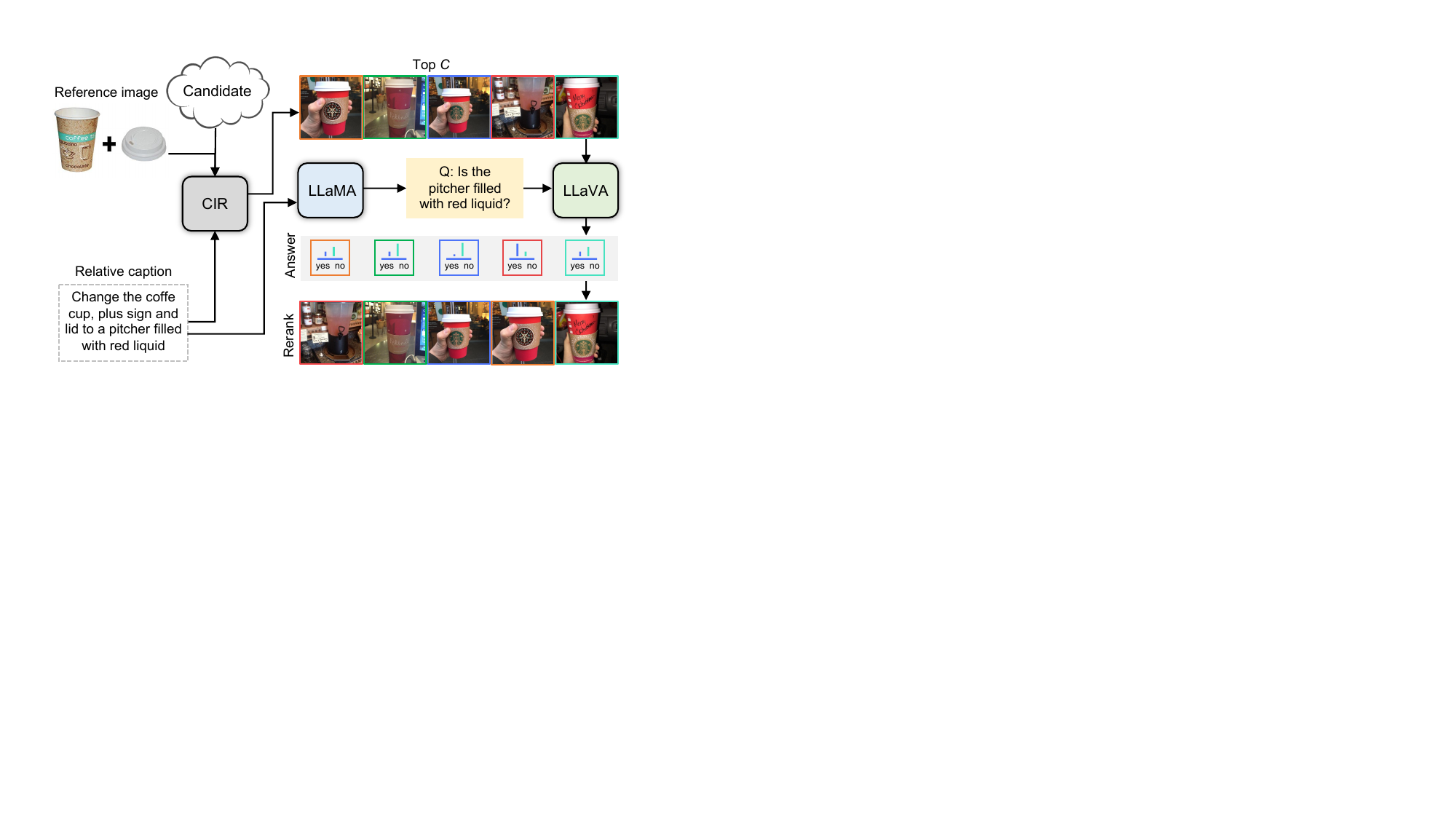}
	\end{center}
    \vspace{-13pt}
	\captionsetup{font=small}
	\caption{\small\textbf{Overview} of the \textit{inference process}. For any given CIR model, one can select its top $C$ retrieved images, and send the relative caption to finetuned LLaMA~\cite{touvron2023llama} for generating QA pairs.
    By feeding the candidate images and generated questions to finetuned LLaVA~\cite{liu2023visual}, we acquire a set of answers, and rerank the candidate images by comparing these answers with the ground truth in QA pairs.}
	\vspace{-22pt}
	\label{fig:4}
\end{figure}

\noindent\textbf{Fine-tune LLaMA.}~
LLaMA~\cite{touvron2023llama} is an open-source language model and thus can be finetuned to enhance performance on specialized tasks.
With the instruction datasets, we are equipped to refine LLaMA~\cite{touvron2023llama}, enabling it to generate QA pairs from relative captions.
As illustrated in Fig~\ref{fig:3} (a), we adopt the handcrafted prompt $\boldsymbol{p}$ from GPT-4 and keep it frozen during finetuning.
%
%

Following~\cite{zhang2023lora}, we leverage LoRA~\cite{hu2021lora} to perform efficient fine-tuning, where the backbone is frozen while only the LoRA module is learnable.
Formally, we have
\begin{equation}
\mathbf{y}_{\texttt{QA}}=\mathcal{F}_{\texttt{LLaMA}}\left(\boldsymbol{p}, \boldsymbol{t}\right),
\end{equation}
where $\mathbf{y}_{\texttt{QA}}$ denotes the generated QA pairs from relative caption $\boldsymbol{t}$. 
Denote by $\mathbf{x}_{\texttt{QA}} = \{(\textbf{Q}_1, \textbf{A}_1), ..., (\textbf{Q}_K, \textbf{A}_K)\}$  the ground-truth of QA pairs.
Supervised instruction tuning can then be adopted to finetune LLaMA for generating QA pairs.

%

%
\subsection{VQA for Boosting CIR}
\noindent\textbf{Training Data.}~
To train the VQA model, we construct a training set by using the top-$C$ (\eg $C= 5$ ) retrieved images $\{\boldsymbol{I}_1, ...,  \boldsymbol{I}_c, ..., \boldsymbol{I}_C\}$ for a CIR method (\eg, SPRC), and generated QA pairs $\mathbf{x}_{\texttt{QA}}$. 
We further introduce an indicator variant $y_c$, where $y_c = 1$ if $\boldsymbol{I}_c$ is the target image, else $y_c = -1$.
Then, the training set for training the VQA method can be represented as  $\{(\boldsymbol{I}_c, {y}_c, \mathbf{x}_{\texttt{QA}})\}$
%

\noindent\textbf{Finetune LLaVA.}~
As shown in Fig.~\ref{fig:3} (b), using LLaVA as an example, we froze the backbone model while leveraging the LoRA trainable~\cite{lai2023lisa}.
LLaVA takes a candidate image $\boldsymbol I_c$ and a question $\textbf{Q}_k$ as the input, and outputs the prediction of the answer $\textbf{A}'_k$.
For finetuning LLaVA, we also predict the probability $p_k$ of $\textbf{A}'_k = \textbf{A}_k$, \ie,
\begin{equation}
p_k=\mathcal{F}_{\texttt{LLaVA}}\left(\boldsymbol I_c, \textbf{Q}_k; \textbf{A}_k\right).
\end{equation}
To finetune LLaMA, the training loss is defined by considering all the QA pairs.
When $\boldsymbol{I}_c$ is the target image (\ie, $y_c = 1$), we require  LLaVA to correctly answer all the questions, \ie, $p_k \simeq 1$. 
In contrast, when $\boldsymbol{I}_c$ is not the target image (\ie, $y_c = - 1$), LLaVA is expected to incorrectly answer at least one question. 
%
%
%
%
Thus, the training loss can be written as
\begin{equation}
-\log \left\{\frac{1-y_c}{2} + y_c \left(p_1\!\cdot \!p_2\!\cdot\!\ldots\!p_K\!\right) \right\}.
\end{equation}
%
%
%
In this manner, we can obtain a VQA model that can also be used to distinguish target images from failure-retrieved images, thereby benefiting the CIR performance.


\subsection{Inference Process}~
\label{sec:inference}
With the fine-tuned LLaMA and LLaVA as the VQA models, we can re-rank the output candidates of any CIR model.
As shown in Fig.~\ref{fig:4}, given the relative caption, finetuned LLaMA is used to generate $K$ QA pairs $\{(\textbf{Q}_1, \textbf{A}_1), ..., (\textbf{Q}_K, \textbf{A}_K)\}$.
Then, the top-$C$ candidate images $\mathcal{I} = \left\{\boldsymbol{I}_{1}, \boldsymbol{I}_{2}, ..., \boldsymbol{I}_{C}\right\}$ are first obtained using a CIR model,
\begin{equation}
\mathcal{I} = \mathcal{F}_{\texttt{CIR}}\left(\boldsymbol I_{r}, \boldsymbol {t} \right).
\end{equation}
%
%
For each question $\textbf{Q}_k$, we use the finetuned LLaVA to predict the probability $p_k$ of  $\textbf{A}'_k =  \textbf{A}_k$.
Then, the product of all $p_k$s are used to indicate the consistency between the retrieved image and relative caption,
%
\begin{equation}
p^{A} = p_1\!\cdot\!p_2\!\cdot\!\ldots\!p_K.
\end{equation}
%

\begin{figure}[!t]
    \vspace{-3pt}
	\begin{center}
		\includegraphics[width=\linewidth]{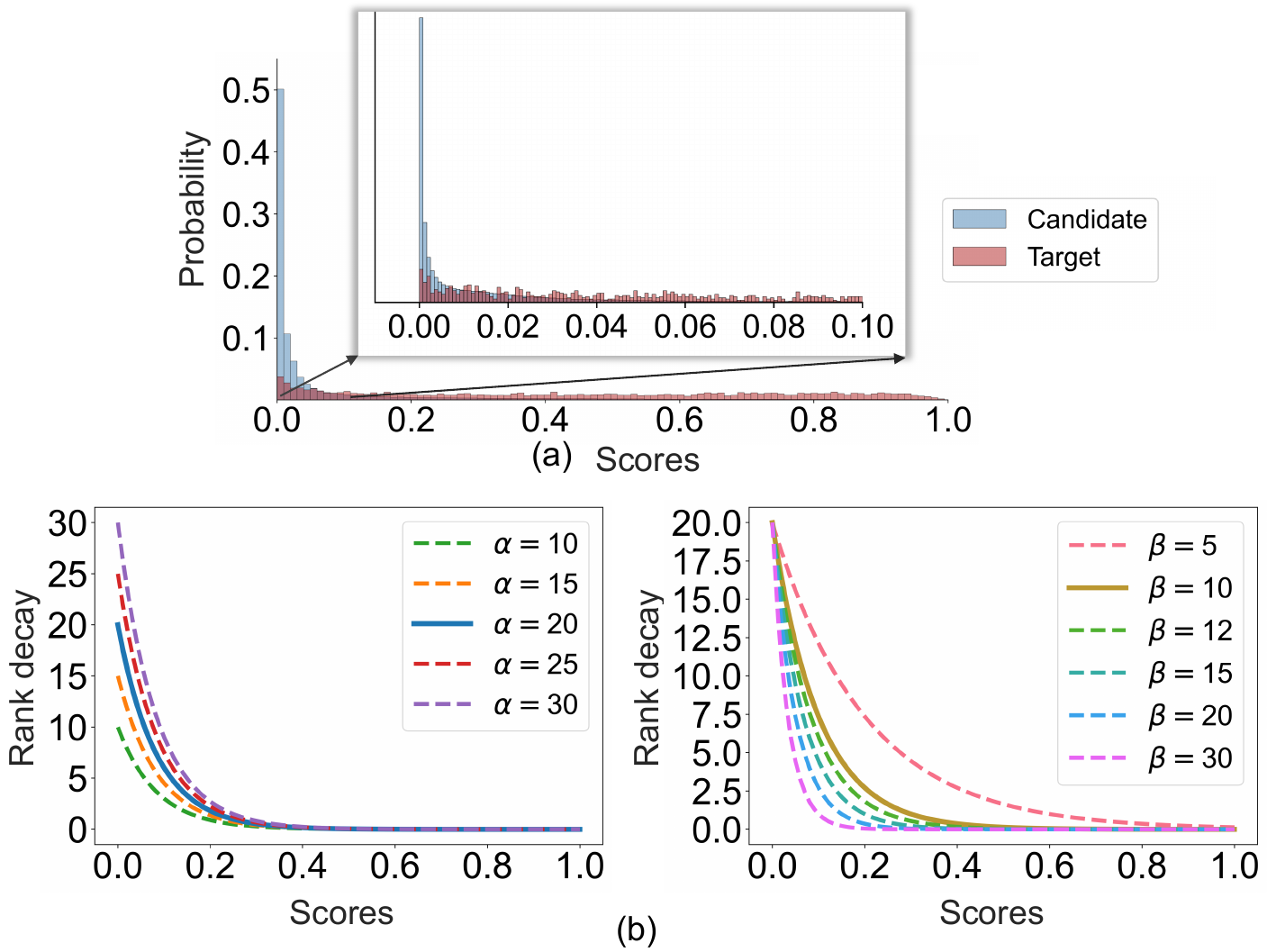}
	\end{center}
    \vspace{-18pt}
	\captionsetup{font=small}
	\caption{\small\textbf{(a) Distribution} visualization of the target predictions and non-target predictions on the CIRR validation sets. \textbf{(b)} Visualization of the \textbf{curves} under different values of $\alpha$ and $\beta$.}
	\vspace{-15pt}
	\label{fig:dis}
\end{figure}

For an ideal VQA model, one can safely reject the candidate images with $p^{A} = 0$, and simply only keep those with $p^{A} = 1$ during reranking.
However, as shown in Fig.~\ref{fig:dis}, on the CIRR validation set, the $p^A$ distributions of target images and failure retrieved images are overlapped. 
%
%
%
%
Thus, we present a soft reranking scheme. 
For a candidate image $\boldsymbol{I}_c$, where its original rank is $c$, we modify its rank to $c+R(p^A)$.
Obviously, $R(p^A)$ should be larger when $p^A$ is small and should be near zero when $p^A \simeq 1$.
To this end, we define $R(p^A)$ as follows, 
%
%
\begin{equation}
R(p^A) = \alpha \mathbf{e}^{-\beta p^A},
\end{equation}
where $\alpha$ is the step size of ranking descent, and $\beta$ is the rate of ranking decline.  A larger $\alpha$ value indicates a greater step size in the descent, and vice versa. A larger $\beta$ indicates a faster rate of decline, refer to Fig.~\ref{fig:dis} (b).
Detailed discussion regarding these two hyperparameters can be seen in Fig.~\ref{fig:alpha}.
Finally, we sort all $c+R(p^A)$ values to give the ranks after reranking.

%% file: sec/4_Experiments.tex
\section{Experiments}

\begin{table*}[t]
\renewcommand{\arraystretch}{1.3} 
	\caption{\small \textbf{Quantitative comparison} in terms of recalls across competing methods on the \texttt{test} set of CIRR, where best and second-best results are highlighted in \textbf{bold} and \underline{underlined}, respectively. The arrow ${\color{ForestGreen}\uparrow}$ indicates improvements compared with baseline models, \ie, CLIP4CIR$^*$~\cite{baldrati2023composed} and SPRC~\cite{bai2023sentence}. Detailed analyses are provided in Sec.~\ref{sec:sota}.}
	\vspace{-7pt}
	\label{tab:2}
        \setlength{\tabcolsep}{0.0pt}
        
	\fontsize{8}{8}\selectfont
	\centering
	\begin{tabular}{l cc cc cc cc cc cc cc cc}
\toprule
\multicolumn{1}{l}{\multirow{2}{*}[-1mm]{\textbf{Method}}}
&\multicolumn{8}{c}{\textbf{\texttt{Recall@K}}}
&\multicolumn{6}{c}{\textbf{\texttt{Recall$_{\text{Subset}}$}}} 
&\multicolumn{2}{c}{\multirow{2}{*}{\textbf{Average}}}
\\

\cmidrule(lr){2-9} \cmidrule(l){10-15}

&\multicolumn{2}{c}{\textbf{K=1}} 
&\multicolumn{2}{c}{\textbf{K=5}} 
&\multicolumn{2}{c}{\textbf{K=10}} 
&\multicolumn{2}{c}{\textbf{K=50}} 
&\multicolumn{2}{c}{\textbf{K=1}} 
&\multicolumn{2}{c}{\textbf{K=2}} 
&\multicolumn{2}{c}{\textbf{K=3}} 
&\multicolumn{2}{c}{\textbf{}} 
\\

\cmidrule(r){1-1} 
\cmidrule{2-3} 
\cmidrule(lr){4-5} 
\cmidrule(lr){6-7} 
\cmidrule(lr){8-9}
\cmidrule(lr){10-11}
\cmidrule(lr){12-13}
\cmidrule(lr){14-15}
\cmidrule(lr){16-17}

TIRG~\cite{vo2019composing}
&\multicolumn{2}{c}{$14.61$} 
&\multicolumn{2}{c}{$48.37$}
&\multicolumn{2}{c}{$64.08$}
&\multicolumn{2}{c}{$90.03$} 
&\multicolumn{2}{|c}{$22.67$}
&\multicolumn{2}{c}{$44.97$}
&\multicolumn{2}{c}{$65.14$}
&\multicolumn{2}{|c}{$35.52$}
\\ 

TIRG+LConv~\cite{vo2019composing}
&\multicolumn{2}{c}{$11.04$} 
&\multicolumn{2}{c}{$35.68$}
&\multicolumn{2}{c}{$51.27$}
&\multicolumn{2}{c}{$83.29$} 
&\multicolumn{2}{|c}{$23.82$}
&\multicolumn{2}{c}{$45.65$}
&\multicolumn{2}{c}{$64.55$}
&\multicolumn{2}{|c}{$29.75$}
\\ 

MAAF~\cite{dodds2020modality}
&\multicolumn{2}{c}{$10.31$} 
&\multicolumn{2}{c}{$33.03$}
&\multicolumn{2}{c}{$48.30$}
&\multicolumn{2}{c}{$80.06$} 
&\multicolumn{2}{|c}{$21.05$}
&\multicolumn{2}{c}{$41.81$}
&\multicolumn{2}{c}{$61.60$}
&\multicolumn{2}{|c}{$27.04$}
\\ 

MAAF-BERT~\cite{dodds2020modality}
&\multicolumn{2}{c}{$10.12$} 
&\multicolumn{2}{c}{$33.10$}
&\multicolumn{2}{c}{$48.01$}
&\multicolumn{2}{c}{$80.57$} 
&\multicolumn{2}{|c}{$22.04$}
&\multicolumn{2}{c}{$42.41$}
&\multicolumn{2}{c}{$62.14$}
&\multicolumn{2}{|c}{$27.57$}
\\ 

MAAF-IT~\cite{dodds2020modality}
&\multicolumn{2}{c}{$9.90$} 
&\multicolumn{2}{c}{$32.86$}
&\multicolumn{2}{c}{$48.83$}
&\multicolumn{2}{c}{$80.27$} 
&\multicolumn{2}{|c}{$21.17$}
&\multicolumn{2}{c}{$42.04$}
&\multicolumn{2}{c}{$60.91$}
&\multicolumn{2}{|c}{$27.02$}
\\ 

MAAF-RP~\cite{dodds2020modality}
&\multicolumn{2}{c}{$10.22$} 
&\multicolumn{2}{c}{$33.32$}
&\multicolumn{2}{c}{$48.68$}
&\multicolumn{2}{c}{$81.84$} 
&\multicolumn{2}{|c}{$21.41$}
&\multicolumn{2}{c}{$42.17$}
&\multicolumn{2}{c}{$61.60$}
&\multicolumn{2}{|c}{$27.37$}
\\ 

CIRPLANT~\cite{liu2021image}
&\multicolumn{2}{c}{$19.55$} 
&\multicolumn{2}{c}{$52.55$}
&\multicolumn{2}{c}{$68.39$}
&\multicolumn{2}{c}{$92.38$} 
&\multicolumn{2}{|c}{$39.20$}
&\multicolumn{2}{c}{$63.03$}
&\multicolumn{2}{c}{$79.49$}
&\multicolumn{2}{|c}{$45.88$}
\\ 


LF-BLIP~\cite{baldrati2022effective}
&\multicolumn{2}{c}{$20.89$} 
&\multicolumn{2}{c}{$48.07$}
&\multicolumn{2}{c}{$61.16$}
&\multicolumn{2}{c}{$83.71$} 
&\multicolumn{2}{|c}{$50.22$}
&\multicolumn{2}{c}{$73.16$}
&\multicolumn{2}{c}{$86.82$}
&\multicolumn{2}{|c}{$60.58$}
\\ 

LF-CLIP~\cite{baldrati2022effective}
&\multicolumn{2}{c}{$33.59$} 
&\multicolumn{2}{c}{$65.35$}
&\multicolumn{2}{c}{$77.35$}
&\multicolumn{2}{c}{$95.21$} 
&\multicolumn{2}{|c}{$62.39$}
&\multicolumn{2}{c}{$81.81$}
&\multicolumn{2}{c}{$92.02$}
&\multicolumn{2}{|c}{$72.53$}
\\

CLIP4CIR~\cite{baldrati2022conditioned}
&\multicolumn{2}{c}{$38.53$} 
&\multicolumn{2}{c}{$69.98$}
&\multicolumn{2}{c}{$81.86$}
&\multicolumn{2}{c}{$95.93$} 
&\multicolumn{2}{|c}{$68.19$}
&\multicolumn{2}{c}{$85.64$}
&\multicolumn{2}{c}{$94.17$}
&\multicolumn{2}{|c}{$69.09$}
\\ 

BLIP4CIR+Bi~\cite{liu2023bi}
&\multicolumn{2}{c}{$40.15$} 
&\multicolumn{2}{c}{$73.08$}
&\multicolumn{2}{c}{$83.88$}
&\multicolumn{2}{c}{$96.27$} 
&\multicolumn{2}{|c}{$72.10$}
&\multicolumn{2}{c}{$88.27$}
&\multicolumn{2}{c}{$95.93$}
&\multicolumn{2}{|c}{$72.59$}
\\ 

CompoDiff~\cite{gu2023compodiff}
&\multicolumn{2}{c}{$22.35$} 
&\multicolumn{2}{c}{$54.36$}
&\multicolumn{2}{c}{$73.41$}
&\multicolumn{2}{c}{$91.77$} 
&\multicolumn{2}{|c}{$35.84$}
&\multicolumn{2}{c}{$56.11$}
&\multicolumn{2}{c}{$76.60$}
&\multicolumn{2}{|c}{$29.10$}
\\ 

CASE~\cite{levy2023data}
&\multicolumn{2}{c}{$48.00$} 
&\multicolumn{2}{c}{$79.11$}
&\multicolumn{2}{c}{$87.2$5}
&\multicolumn{2}{c}{$97.57$} 
&\multicolumn{2}{|c}{$75.88$}
&\multicolumn{2}{c}{$90.58$}
&\multicolumn{2}{c}{$96.00$}
&\multicolumn{2}{|c}{$77.50$}
\\


TG-CIR~\cite{wen2023target}
&\multicolumn{2}{c}{$45.25$} 
&\multicolumn{2}{c}{$78.29$}
&\multicolumn{2}{c}{$87.16$}
&\multicolumn{2}{c}{$97.30$} 
&\multicolumn{2}{|c}{$72.84$}
&\multicolumn{2}{c}{$89.25$}
&\multicolumn{2}{c}{$95.13$}
&\multicolumn{2}{|c}{$75.57$}
\\ 

DRA~\cite{jiang2023dual}
&\multicolumn{2}{c}{$39.93$} 
&\multicolumn{2}{c}{$72.07$}
&\multicolumn{2}{c}{$83.83$}
&\multicolumn{2}{c}{$96.43$} 
&\multicolumn{2}{|c}{$71.04$}
&\multicolumn{2}{c}{$87.74$}
&\multicolumn{2}{c}{$94.72$}
&\multicolumn{2}{|c}{$71.55$}
\\ 

CoVR-BLIP~\cite{ventura2023covr}
&\multicolumn{2}{c}{$49.69$} 
&\multicolumn{2}{c}{$78.60$}
&\multicolumn{2}{c}{$86.77$}
&\multicolumn{2}{c}{$94.31$} 
&\multicolumn{2}{|c}{$75.01$}
&\multicolumn{2}{c}{$88.12$}
&\multicolumn{2}{c}{$93.16$}
&\multicolumn{2}{|c}{$80.81$}
\\ 

Re-ranking~\cite{liu2023candidate}
&\multicolumn{2}{c}{{$50.55$}} 
&\multicolumn{2}{c}{{$81.75$}}
&\multicolumn{2}{c}{{$89.78$}}
&\multicolumn{2}{c}{{$97.18$}} 
&\multicolumn{2}{|c}{{$80.04$}}
&\multicolumn{2}{c}{{$91.90$}}
&\multicolumn{2}{c}{{$96.58$}}
&\multicolumn{2}{|c}{{$80.90$}}
\\

\cmidrule(r){1-1} 
\cmidrule(lr){2-9} 
\cmidrule(lr){10-15} 
\cmidrule(lr){16-17}


{\cellcolor{mygray}CLIP4CIR$^*$~\cite{baldrati2023composed}}
&\multicolumn{2}{l}{{\cellcolor{mygray}\qquad{$44.82$}}} 
&\multicolumn{2}{l}{{\cellcolor{mygray}\qquad{$77.04$}}} 
&\multicolumn{2}{l}{{\cellcolor{mygray}\qquad{$86.65$}}}
&\multicolumn{2}{l}{{\cellcolor{mygray}\qquad{$97.90$}}}
&\multicolumn{2}{|l}{{\cellcolor{mygray}\qquad{$73.16$}}} 
&\multicolumn{2}{l}{{\cellcolor{mygray}\qquad{$88.84$}}}
&\multicolumn{2}{l}{{\cellcolor{mygray}\qquad{$95.59$}}}
&\multicolumn{2}{|l}{{\cellcolor{mygray}\qquad{$75.10$}}}\\


{\cellcolor{myblue}+ \textbf{\texttt{VQA4CIR}}}
&\multicolumn{2}{l}{{\cellcolor{myblue}\qquad{$51.40$}}\stdvu{${{6.6}}$}} 
&\multicolumn{2}{l}{{\cellcolor{myblue}\qquad{$81.71$}}\stdvu{${{4.7}}$}} 
&\multicolumn{2}{l}{{\cellcolor{myblue}\qquad{\underline{$89.83$}}}\stdvu{${{3.2}}$}}
&\multicolumn{2}{l}{{\cellcolor{myblue}\qquad\underline{$98.10$}}\stdvu{${{0.2}}$}}
&\multicolumn{2}{|l}{{\cellcolor{myblue}\qquad{$80.15$}}\stdvu{${{7.0}}$}} 
&\multicolumn{2}{l}{{\cellcolor{myblue}\qquad\underline{$92.63$}}\stdvu{${{3.8}}$}}
&\multicolumn{2}{l}{{\cellcolor{myblue}\qquad{\underline{$96.75$}}}\stdvu{${{1.1}}$}}
&\multicolumn{2}{|l}{{\cellcolor{myblue}\qquad{{$80.93$}}}\stdvu{${{5.8}}$}}\\


{\cellcolor{mygray}SPRC~\cite{bai2023sentence}}
&\multicolumn{2}{c}{{\cellcolor{mygray}\underline{$51.96$}}} 
&\multicolumn{2}{c}{{\cellcolor{mygray}\underline{$82.12$}}} 
&\multicolumn{2}{c}{{\cellcolor{mygray}$89.74$}}
&\multicolumn{2}{c}{{\cellcolor{mygray}{$97.69$}}}
&\multicolumn{2}{|c}{{\cellcolor{mygray}\underline{$80.65$}}} 
&\multicolumn{2}{c}{{\cellcolor{mygray}{$92.31$}}}
&\multicolumn{2}{c}{{\cellcolor{mygray}$96.60$}}
&\multicolumn{2}{|c}{{\cellcolor{mygray}\underline{$81.39$}}}\\


{\cellcolor{myblue}+ \textbf{\texttt{VQA4CIR}}}
&\multicolumn{2}{l}{{\cellcolor{myblue}\qquad\textbf{54.00}}\stdvu{${{2.0}}$}} 
&\multicolumn{2}{l}{{\cellcolor{myblue}\qquad\textbf{84.23}}\stdvu{${{2.1}}$}} 
&\multicolumn{2}{l}{{\cellcolor{myblue}\qquad\textbf{91.85}}\stdvu{${{2.1}}$}}
&\multicolumn{2}{l}{{\cellcolor{myblue}\qquad\textbf{98.10}}\stdvu{${{0.4}}$}}
&\multicolumn{2}{|l}{{\cellcolor{myblue}\qquad\textbf{82.07}}\stdvu{${{1.4}}$}}
&\multicolumn{2}{l}{{\cellcolor{myblue}\qquad\textbf{93.45}}\stdvu{${{1.1}}$}}
&\multicolumn{2}{l}{{\cellcolor{myblue}\qquad\textbf{97.08}}\stdvu{${{0.5}}$}}
&\multicolumn{2}{|l}{{\cellcolor{myblue}\qquad\textbf{83.15}}\stdvu{${{1.8}}$}}\\

\bottomrule
\end{tabular}
\vspace{-12pt}
\end{table*}

\subsection{Experimental Setup}

\noindent{\textbf{Implementation Details.}}
Our VQA4CIR is implemented with Pytorch on NVIDIA RTX A100 GPUs with $40$GB of memory per card.
To preserve the generalization ability of the pre-trained models, \ie, LLaMA~\cite{touvron2023llama} and LLaVA~\cite{liu2023visual}, we leverage LoRA~\cite{hu2021lora} to fine-tune them while keeping the backbones frozen, \ie, LLaVA-v1.5-13B and Vicuna-13B-v1.5.
We note the word embeddings of the LLaMA~\cite{touvron2023llama} and LLaVA~\cite{liu2023visual} are also frozen.
%
During training, we randomly adopt $5,000$ and $3,000$ samples from the CIRR dataset and Fashion-IQ training data, respectively, to fine-tune LLaMA~\cite{touvron2023llama} and LLaVA~\cite{liu2023visual}.
The AdamW~\cite{loshchilov2017decoupled} is adopted
as the optimizer with a weight decay of $0.05$ across all the experiments.
We adopt WarmupDecayLR as the learning rate scheduler with warmup iterations of $1,000$.
%
For LLaVA~\cite{liu2023visual}, the learning rate is initialized at $2$e-$5$, while for LLaMA~\cite{touvron2023llama}, it is initialized at $3$e-$4$. 
The hyperparameter of $\alpha$ is respectively set to $20$ and $30$ on the CIRR and Fashion-IQ datasets, while $\beta$ is empirically set to $10$ and $12$.

\noindent{\textbf{Datasets.}}
We evaluate our method on two CIR benchmarks: \texttt{(1)} Fashion-IQ si a fashion dataset, which consists of three fashion categories, \ie, Shirt, Dress, and Toptee, with $77,684$ images forming $30,134$ triplets~\cite{wu2021fashion}. 
According to the standard experimental setup of this dataset, we employ the Recall@K evaluation metric, representing the proportion of queries where the actual target is ranked among the top $K$ candidates.
In our experiments, we report the results using two distinct levels of recall, \eg, Recall@$10$ and Recall@$50$, for each category.
Given that the ground-truth labels for the test set of this dataset remain undisclosed, we assess performance using the validation set.
%
%
\texttt{(2)} CIRR is another CIR dataset consisting of $36,554$ triplets, sourced from $21,552$ images in the widely-recognized natural language inference dataset, \ie, NLVR2~\cite{suhr2018corpus}.
Following the standard experimental setup of this dataset, we split this dataset into \texttt{training}, \texttt{validation}, and \texttt{test} sets with a ratio of $8:1:1$. 
In contrast to Fashion-IQ, this dataset offers a wide range of object interactions and variance, and may be more suitable for assessing the effectiveness of our proposed method.
%
%
%
We showcase the performance of the state-of-the-art methods on this dataset using the metrics Recall@$1$, $5$, $10$, $50$, and Recall$_{\text{Subset}}$~\cite{liu2021image}.
%

\subsection{Comparison with State-of-the-arts}\label{sec:sota}

\noindent{\textbf{CIRR.}}~In Table~\ref{tab:2} lists the results on the CIRR dataset.
To evaluate whether VQA can assist CIR, we selected the most classical method, CLIP4CIR$^*$~\cite{baldrati2023composed}, and the currently best-performing method, SPRC~\cite{bai2023sentence}, as our CIR base models.
As can be seen, our method achieves noticeable improvements across all the metrics.
%
Although the baseline SPRC has the highest performance among the competing methods, our VQA4CIR can further improve its recall values, \eg, Recall@K=10, +VQA4CIR: \textbf{91.85}, SPRC~\cite{bai2023sentence}: $89.74$.
%
%
%
Benefited from VQA4CIR, our method by adopting CLIP4CIR as the base model, achieves the second-best performance under Recall@K=10 and RecallSubset@K=2, \ie, $86.65$ $\rightarrow$ $\textbf{89.83}$, $88.84$ $\rightarrow$ $\textbf{92.63}$, referring to the underlined results.
Compared to the re-ranking method~\cite{liu2023candidate}, our method attains a notable improvement on all metrics, \eg, Recall@K=1, $50.55$ $\rightarrow$ $\textbf{54.00}$, Avg. $80.90$ $\rightarrow$ $\textbf{83.15}$.
%
%
%

\noindent{\textbf{Fashion-IQ.}}~
%
Table~\ref{tab:1} lists the results of competing methods on the Fashion-IQ dataset.
As can be seen, our VQA4CIR achieves consistent performance gains across eight evaluation metrics in three different categories, \ie, it has the highest recall values, as indicated by the bold results in the table.
Although SPRC~\cite{bai2023sentence} has the highest performance among all the existing baseline methods, our method can further improve its recalls, \eg, in class \texttt{\textbf{Dress}}, R@10: $47.80$ to $\textbf{49.18}$, and R@50: $72.70$ to $\textbf{73.06}$, and there is also a significant improvement in the average recall, \ie, $64.76$ \textit{\textbf{vs.}} $\textbf{65.41}$.
For CLIP4CIR~\cite{baldrati2022conditioned}, we used its enhanced version, CLIP4CIR$^*$~\cite{baldrati2023composed}.
By incorporating VQA4CIR with CLIP4CIR, our method has a significant improvement, \eg, in class \texttt{\textbf{Dress}}, R@10: $39.46$ to $\textbf{40.91}$, and R@50: $64.55$ to $\textbf{65.13}$, and there is also a significant improvement in the average recalls, \ie, $55.35$ \textit{\textbf{vs.}} $\textbf{56.29}$.
Moreover, the re-ranking method~\cite{liu2023candidate} still falls below ours, \eg, in average recalls, $62.15$ \textit{\textbf{vs.}} $\textbf{65.41}$.
%
%
Although VQA4CIR also involves a stage of re-ranking, it is essentially different from~\cite{liu2023candidate}. 
%
%
%
Refer to \textit{\textbf{Suppl}.} for the qualitative evaluation of the two datasets.

\subsection{Ablation Studies}\label{sec:ab}

\noindent{\textbf{Top $C$ for Re-ranking.}}~
Our method involves re-ranking the output of existing CIR models, 
%
and we thus discuss the effect of the number of the top-$C$ retrieved images by the base CIR method, \ie, the effect of $C$ on CIR performance.
%
Using the CIRR dataset, we provide the results of different $C$ values on the validation sets of CIRR in Table~\ref{tab:3}.
One can see that as $C$ increases, there is an improvement in recall, possibly ascribing to that the larger $C$ is, the greater the coverage of ground truth, thereby yielding higher performance.
Nonetheless, after reaching a peak, further increasing $C$ leads to more negative candidate images, which may affect performance if the VQA model cannot correctly find all inconsistent retrieved images, \eg, Top $150$ yielding worse results than Top $15$.
It is noteworthy that our method achieves the highest improvement when $C=15$.
The smaller the value of $C$, the more pronounced the improvement, \eg, under the Recall@K metric with $C = 1$, recalls improve from top $0$: $53.94$ to top $15$: \textbf{56.15}, and under the Recall$_{\text{Subset}}$ metric with $C = 1$, recalls improve from top $0$: $79.78$ to top $15$: \textbf{82.56}.
%
Here, we note that $C$ also improves the validation of the Recall$_{\text{Subset}}$ while such a metric only involves five predetermined candidates for each query.
Our method achieves a noticeable improvement on this metric, indicating that the VQA mechanism can effectively aid the model in recognizing the candidate image being inconsistent with relative caption. 
In contrast, Liu \textit{et al}. shows almost no change on this metric~\cite{liu2023candidate}.
We also note that the inference time cost is positively correlated with the value of top $C$.
In this paper, considering the results in Table~\ref{tab:3}, as well as the inference costs, we choose top $70$ across all the experiments.

\begin{table*}[t]
\vspace{-14pt}
\renewcommand{\arraystretch}{1.3} 
\setlength{\tabcolsep}{5pt} 
	\caption{\small \textbf{Quantitative comparison} in terms of {recalls} of various methods on the validation set of the Fashion-IQ dataset. Best and second-best results are highlighted in \textbf{bold} and \underline{underlined}, respectively. $^*$ indicates an improved version. The arrow ${\color{ForestGreen}\uparrow}$ indicates improvements compared with baseline models, \ie, CLIP4CIR$^*$~\cite{baldrati2023composed} and SPRC~\cite{bai2023sentence}. Detailed analyses are provided in Sec.~\ref{sec:sota}.}
	\vspace{-8pt}
	\label{tab:1}

	\fontsize{8}{8}\selectfont
	\centering
	\begin{tabular}{l cc cc cc cc cc cc cc cc cc cc}
\toprule
\multicolumn{1}{l}{\multirow{2}{*}[-1mm]{\textbf{Method}}}
&\multicolumn{4}{c}{\textbf{\texttt{Dress}}}
&\multicolumn{4}{c}{\textbf{\texttt{Shirt}}} 
&\multicolumn{4}{c}{\textbf{\texttt{Toptee}}} 
&\multicolumn{4}{c}{\textbf{\texttt{Average}}} 
&\multicolumn{2}{c}{\multirow{2}{*}{\textbf{Avg.}}}
\\

\cmidrule(lr){2-5} \cmidrule(lr){6-9}\cmidrule(lr){10-13}\cmidrule(lr){14-17}

&\multicolumn{2}{c}{\textbf{R@10}} 
&\multicolumn{2}{c}{\textbf{R@50}} 
&\multicolumn{2}{c}{\textbf{R@10}} 
&\multicolumn{2}{c}{\textbf{R@50}} 
&\multicolumn{2}{c}{\textbf{R@10}} 
&\multicolumn{2}{c}{\textbf{R@50}} 
&\multicolumn{2}{c}{\textbf{R@10}} 
&\multicolumn{2}{c}{\textbf{R@50}} 
&
\\

\cmidrule(r){1-1} 
\cmidrule{2-3} 
\cmidrule(lr){4-5} 
\cmidrule(lr){6-7} 
\cmidrule(lr){8-9}
\cmidrule(lr){10-11}
\cmidrule(lr){12-13}
\cmidrule(lr){14-15}
\cmidrule(lr){16-17}
\cmidrule(lr){18-19}

JVSM~\cite{chen2020learning}  
&\multicolumn{2}{c}{$10.70$} 
&\multicolumn{2}{c}{$25.90$}
&\multicolumn{2}{|c}{$12.00$}
&\multicolumn{2}{c}{$27.10$} 
&\multicolumn{2}{|c}{$13.00$}
&\multicolumn{2}{c}{$26.90$}
&\multicolumn{2}{|c}{$11.90$}
&\multicolumn{2}{c}{$26.60$}
&\multicolumn{2}{|c}{$19.26$}
\\

CIRPLANT~\cite{liu2021image}
&\multicolumn{2}{c}{$17.45$} 
&\multicolumn{2}{c}{$40.41$}
&\multicolumn{2}{|c}{$17.53$}
&\multicolumn{2}{c}{$38.81$} 
&\multicolumn{2}{|c}{$61.64$}
&\multicolumn{2}{c}{$45.38$}
&\multicolumn{2}{|c}{$18.87$}
&\multicolumn{2}{c}{$41.53$}
&\multicolumn{2}{|c}{$30.20$}
\\ 


VAL w/GloVe~\cite{chen2020image}
&\multicolumn{2}{c}{$22.53$} 
&\multicolumn{2}{c}{$44.00$}
&\multicolumn{2}{|c}{$22.38$}
&\multicolumn{2}{c}{$44.15$} 
&\multicolumn{2}{|c}{$27.53$}
&\multicolumn{2}{c}{$51.68$}
&\multicolumn{2}{|c}{$24.15$}
&\multicolumn{2}{c}{$46.61$}
&\multicolumn{2}{|c}{$35.38$}
\\ 




CoSMo~\cite{lee2021cosmo}
&\multicolumn{2}{c}{$25.64$} 
&\multicolumn{2}{c}{$50.30$}
&\multicolumn{2}{|c}{$24.90$}
&\multicolumn{2}{c}{$49.18$} 
&\multicolumn{2}{|c}{$29.21$}
&\multicolumn{2}{c}{$57.46$}
&\multicolumn{2}{|c}{$26.58$}
&\multicolumn{2}{c}{$52.31$}
&\multicolumn{2}{|c}{$39.45$}
\\ 


DCNet~\cite{kim2021dual}
&\multicolumn{2}{c}{$28.95$} 
&\multicolumn{2}{c}{$56.07$}
&\multicolumn{2}{|c}{$23.95$}
&\multicolumn{2}{c}{$47.30$} 
&\multicolumn{2}{|c}{$30.44$}
&\multicolumn{2}{c}{$58.29$}
&\multicolumn{2}{|c}{$27.78$}
&\multicolumn{2}{c}{$53.89$}
&\multicolumn{2}{|c}{$40.84$}
\\ 

SAC w/BERT~\cite{jandial2022sac}
&\multicolumn{2}{c}{$26.52$} 
&\multicolumn{2}{c}{$51.01$}
&\multicolumn{2}{|c}{$28.02$}
&\multicolumn{2}{c}{$51.86$} 
&\multicolumn{2}{|c}{$32.70$}
&\multicolumn{2}{c}{$61.23$}
&\multicolumn{2}{|c}{$29.08$}
&\multicolumn{2}{c}{$54.70$}
&\multicolumn{2}{|c}{$41.89$}
\\ 

FashionVLP~\cite{goenka2022fashionvlp}
&\multicolumn{2}{c}{$32.42$} 
&\multicolumn{2}{c}{$60.29$}
&\multicolumn{2}{|c}{$31.89$}
&\multicolumn{2}{c}{$58.44$} 
&\multicolumn{2}{|c}{$38.51$}
&\multicolumn{2}{c}{$68.79$}
&\multicolumn{2}{|c}{$34.27$}
&\multicolumn{2}{c}{$62.51$}
&\multicolumn{2}{|c}{$48.39$}
\\ 

LF-CLIP~\cite{baldrati2022effective}
&\multicolumn{2}{c}{$31.63$} 
&\multicolumn{2}{c}{$56.67$}
&\multicolumn{2}{|c}{$36.36$}
&\multicolumn{2}{c}{$58.00$} 
&\multicolumn{2}{|c}{$38.19$}
&\multicolumn{2}{c}{$62.42$}
&\multicolumn{2}{|c}{$35.39$}
&\multicolumn{2}{c}{$59.03$}
&\multicolumn{2}{|c}{$47.21$}
\\ 

LF-BLIP~\cite{baldrati2022effective}
&\multicolumn{2}{c}{$25.31$} 
&\multicolumn{2}{c}{$44.05$}
&\multicolumn{2}{|c}{$25.39$}
&\multicolumn{2}{c}{$43.57$} 
&\multicolumn{2}{|c}{$26.54$}
&\multicolumn{2}{c}{$44.48$}
&\multicolumn{2}{|c}{$25.75$}
&\multicolumn{2}{c}{$43.98$}
&\multicolumn{2}{|c}{$34.88$}
\\ 

CASE~\cite{levy2023data}
&\multicolumn{2}{c}{$47.44$} 
&\multicolumn{2}{c}{$69.36$}
&\multicolumn{2}{|c}{$48.48$}
&\multicolumn{2}{c}{$70.23$} 
&\multicolumn{2}{|c}{$50.18$}
&\multicolumn{2}{c}{$72.24$}
&\multicolumn{2}{|c}{$48.79$}
&\multicolumn{2}{c}{$70.68$}
&\multicolumn{2}{|c}{$59.74$}
\\ 

AMC~\cite{zhu2023amc}
&\multicolumn{2}{c}{$31.73$} 
&\multicolumn{2}{c}{$59.25$}
&\multicolumn{2}{|c}{$30.67$}
&\multicolumn{2}{c}{$59.08$} 
&\multicolumn{2}{|c}{$36.21$}
&\multicolumn{2}{c}{$66.06$}
&\multicolumn{2}{|c}{$32.87$}
&\multicolumn{2}{c}{$61.64$}
&\multicolumn{2}{|c}{$47.25$}
\\ 

CoVR-BLIP~\cite{ventura2023covr}
&\multicolumn{2}{c}{$44.55$} 
&\multicolumn{2}{c}{$69.03$}
&\multicolumn{2}{|c}{$48.43$}
&\multicolumn{2}{c}{$67.42$} 
&\multicolumn{2}{|c}{$52.60$}
&\multicolumn{2}{c}{$74.31$}
&\multicolumn{2}{|c}{$48.53$}
&\multicolumn{2}{c}{$70.25$}
&\multicolumn{2}{|c}{$59.39$}
\\ 

CLIP4CIR~\cite{baldrati2022conditioned}
&\multicolumn{2}{c}{$33.81$} 
&\multicolumn{2}{c}{$59.40$}
&\multicolumn{2}{|c}{$39.99$}
&\multicolumn{2}{c}{$60.45$} 
&\multicolumn{2}{|c}{$41.41$}
&\multicolumn{2}{c}{$65.37$}
&\multicolumn{2}{|c}{$38.32$}
&\multicolumn{2}{c}{$61.74$}
&\multicolumn{2}{|c}{$50.03$}
\\

BLIP4CIR+Bi~\cite{liu2023bi}
&\multicolumn{2}{c}{$42.09$} 
&\multicolumn{2}{c}{$67.33$}
&\multicolumn{2}{|c}{$41.76$}
&\multicolumn{2}{c}{$64.28$} 
&\multicolumn{2}{|c}{$46.61$}
&\multicolumn{2}{c}{$70.32$}
&\multicolumn{2}{|c}{$43.49$}
&\multicolumn{2}{c}{$67.31$}
&\multicolumn{2}{|c}{$55.04$}
\\ 

FAME-ViL$^\dagger$~\cite{han2023fame}
&\multicolumn{2}{c}{$42.19$} 
&\multicolumn{2}{c}{$67.38$}
&\multicolumn{2}{|c}{$47.64$}
&\multicolumn{2}{c}{$68.79$} 
&\multicolumn{2}{|c}{$50.69$}
&\multicolumn{2}{c}{$73.07$}
&\multicolumn{2}{|c}{$46.84$}
&\multicolumn{2}{c}{$69.75$}
&\multicolumn{2}{|c}{$58.29$}
\\ 

TG-CIR~\cite{wen2023target}
&\multicolumn{2}{c}{$45.22$} 
&\multicolumn{2}{c}{$69.66$}
&\multicolumn{2}{|c}{$52.6$0}
&\multicolumn{2}{c}{$72.52$} 
&\multicolumn{2}{|c}{$56.14$}
&\multicolumn{2}{c}{$77.10$}
&\multicolumn{2}{|c}{$51.32$}
&\multicolumn{2}{c}{$73.09$}
&\multicolumn{2}{|c}{$58.05$}
\\ 

DRA~\cite{jiang2023dual}
&\multicolumn{2}{c}{$33.98$} 
&\multicolumn{2}{c}{$60.67$}
&\multicolumn{2}{|c}{$40.74$}
&\multicolumn{2}{c}{$61.93$} 
&\multicolumn{2}{|c}{$42.09$}
&\multicolumn{2}{c}{$66.97$}
&\multicolumn{2}{|c}{$38.93$}
&\multicolumn{2}{c}{$63.19$}
&\multicolumn{2}{|c}{$51.06$}
\\ 

Re-ranking~\cite{liu2023candidate}
&\multicolumn{2}{c}{\underline{$48.14$}} 
&\multicolumn{2}{c}{{$71.43$}}
&\multicolumn{2}{|c}{{$50.15$}}
&\multicolumn{2}{c}{{$71.25$}} 
&\multicolumn{2}{|c}{{$55.23$}}
&\multicolumn{2}{c}{{$76.80$}}
&\multicolumn{2}{|c}{{$51.17$}}
&\multicolumn{2}{c}{{$73.13$}}
&\multicolumn{2}{|c}{{$62.15$}}
\\

CompoDiff~\cite{gu2023compodiff}
&\multicolumn{2}{c}{$40.65$} 
&\multicolumn{2}{c}{$57.14$}
&\multicolumn{2}{|c}{$36.87$}
&\multicolumn{2}{c}{$57.39$} 
&\multicolumn{2}{|c}{$43.93$}
&\multicolumn{2}{c}{$61.17$}
&\multicolumn{2}{|c}{$40.48$}
&\multicolumn{2}{c}{$58.57$}
&\multicolumn{2}{|c}{$49.53$}
\\

\cmidrule(r){1-1} 
\cmidrule(lr){2-5} 
\cmidrule(lr){6-9} 
\cmidrule(lr){10-13} 
\cmidrule(lr){14-17} 
\cmidrule(lr){18-19}

{\cellcolor{mygray}CLIP4CIR$^*$~\cite{baldrati2023composed}}
&\multicolumn{2}{c}{{\cellcolor{mygray}{$39.46$}}} 
&\multicolumn{2}{c}{{\cellcolor{mygray}{$64.55$}}} 
&\multicolumn{2}{|c}{{\cellcolor{mygray}{$44.41$}}}
&\multicolumn{2}{c}{{\cellcolor{mygray}{$65.26$}}}
&\multicolumn{2}{|c}{{\cellcolor{mygray}{$47.48$}}} 
&\multicolumn{2}{c}{{\cellcolor{mygray}{$70.98$}}}
&\multicolumn{2}{|c}{{\cellcolor{mygray}{$43.78$}}}
&\multicolumn{2}{c}{{\cellcolor{mygray}{$66.93$}}}
&\multicolumn{2}{|c}{{\cellcolor{mygray}{$55.35$}}}\\

{\cellcolor{myblue}+ \textbf{\texttt{VQA4CIR}}}
&\multicolumn{2}{c}{{\cellcolor{myblue}{$40.91$}}\stdvu{${{1.5}}$}}
&\multicolumn{2}{c}{{\cellcolor{myblue}{$65.13$}}\stdvu{${{0.6}}$}}
&\multicolumn{2}{|c}{{\cellcolor{myblue}{$45.62$}}\stdvu{${{1.2}}$}} 
&\multicolumn{2}{c}{{\cellcolor{myblue}{$65.68$}}\stdvu{${{0.4}}$}} 
&\multicolumn{2}{|c}{{\cellcolor{myblue}{$49.21$}}\stdvu{${{1.7}}$}} 
&\multicolumn{2}{c}{{\cellcolor{myblue}{$71.22$}}\stdvu{${{0.2}}$}}
&\multicolumn{2}{|c}{{\cellcolor{myblue}{$45.24$}}\stdvu{${{1.5}}$}}
&\multicolumn{2}{c}{{\cellcolor{myblue}{$67.34$}}\stdvu{${{0.4}}$}}
&\multicolumn{2}{|c}{{\cellcolor{myblue}{$56.29$}}\stdvu{${{0.9}}$}}\\

{\cellcolor{mygray}SPRC~\cite{bai2023sentence}}
&\multicolumn{2}{c}{{\cellcolor{mygray}{$47.80$}}} 
&\multicolumn{2}{c}{{\cellcolor{mygray}\underline{$72.70$}}} 
&\multicolumn{2}{|c}{{\cellcolor{mygray}\underline{$55.84$}}}
&\multicolumn{2}{c}{{\cellcolor{mygray}\underline{$74.37$}}}
&\multicolumn{2}{|c}{{\cellcolor{mygray}\underline{$58.89$}}} 
&\multicolumn{2}{c}{{\cellcolor{mygray}\underline{$78.99$}}}
&\multicolumn{2}{|c}{{\cellcolor{mygray}\underline{$54.17$}}}
&\multicolumn{2}{c}{{\cellcolor{mygray}\underline{$75.35$}}}
&\multicolumn{2}{|c}{{\cellcolor{mygray}\underline{$64.76$}}}\\

{\cellcolor{myblue}+ \textbf{\texttt{VQA4CIR}}}
&\multicolumn{2}{c}{{\cellcolor{myblue}\textbf{49.18}}\stdvu{${{1.4}}$}} 
&\multicolumn{2}{c}{{\cellcolor{myblue}\textbf{73.06}}\stdvu{${{0.5}}$}} 
&\multicolumn{2}{|c}{{\cellcolor{myblue}\textbf{56.79}}\stdvu{${{1.0}}$}}
&\multicolumn{2}{c}{{\cellcolor{myblue}\textbf{74.52}}\stdvu{${{0.2}}$}}
&\multicolumn{2}{|c}{{\cellcolor{myblue}\textbf{59.67}}\stdvu{${{0.8}}$}} 
&\multicolumn{2}{c}{{\cellcolor{myblue}\textbf{79.30}}\stdvu{${{0.3}}$}}
&\multicolumn{2}{|c}{{\cellcolor{myblue}\textbf{55.21}}\stdvu{${{1.0}}$}}
&\multicolumn{2}{c}{{\cellcolor{myblue}\textbf{75.62}}\stdvu{${{0.3}}$}}
&\multicolumn{2}{|c}{{\cellcolor{myblue}\textbf{65.41}}\stdvu{${{0.7}}$}}\\

\bottomrule
\end{tabular}
\end{table*}

\begin{table*}[!t]
\vspace{-6pt}
\renewcommand{\arraystretch}{1.3} 
\setlength{\tabcolsep}{15pt} 
	\caption{\small \textbf{Ablation studies} in terms of {recalls} with regard to \textit{different Top $C$ values} on the validation set of CIRR dataset.}
	\vspace{-8pt}
	\label{tab:3}

	\fontsize{8}{8}\selectfont
	\centering
	\begin{tabular}{l cc cc cc cc cc cc cc cc}
\toprule
\multicolumn{1}{l}{\multirow{2}{*}[-1mm]{\textbf{Method}}}
&\multicolumn{8}{c}{\textbf{\texttt{Recall@K}}}
&\multicolumn{6}{c}{\textbf{\texttt{Recall$_{\text{Subset}}$}}} 
&\multicolumn{2}{c}{\multirow{2}{*}{\textbf{\texttt{Average}}}}
\\

\cmidrule(lr){2-9} \cmidrule(l){10-15}

&\multicolumn{2}{c}{\textbf{K=1}} 
&\multicolumn{2}{c}{\textbf{K=5}} 
&\multicolumn{2}{c}{\textbf{K=10}} 
&\multicolumn{2}{c}{\textbf{K=50}} 
&\multicolumn{2}{c}{\textbf{K=1}} 
&\multicolumn{2}{c}{\textbf{K=2}} 
&\multicolumn{2}{c}{\textbf{K=3}} 
&\multicolumn{2}{c}{\textbf{}} 
\\

\cmidrule(r){1-1} 
\cmidrule{2-3} 
\cmidrule(lr){4-5} 
\cmidrule(lr){6-7} 
\cmidrule(lr){8-9}
\cmidrule(lr){10-11}
\cmidrule(lr){12-13}
\cmidrule(lr){14-15}
\cmidrule(lr){16-17}

Top $0$ (Baseline)
&\multicolumn{2}{c}{$53.94$} 
&\multicolumn{2}{c}{$84.33$}
&\multicolumn{2}{c}{$90.91$}
&\multicolumn{2}{c}{$98.13$} 
&\multicolumn{2}{|c}{$79.78$}
&\multicolumn{2}{c}{$92.46$}
&\multicolumn{2}{c}{$96.89$}
&\multicolumn{2}{|c}{$82.06$}
\\ 

Top $15$
&\multicolumn{2}{c}{$56.15$} 
&\multicolumn{2}{c}{$85.24$}
&\multicolumn{2}{c}{$92.01$}
&\multicolumn{2}{c}{$98.13$} 
&\multicolumn{2}{|c}{$82.56$}
&\multicolumn{2}{c}{$93.27$}
&\multicolumn{2}{c}{$97.27$}
&\multicolumn{2}{|c}{$83.90$}
\\ 

Top $30$
&\multicolumn{2}{c}{$56.04$} 
&\multicolumn{2}{c}{$85.31$}
&\multicolumn{2}{c}{$92.18$}
&\multicolumn{2}{c}{$98.13$} 
&\multicolumn{2}{|c}{$82.92$}
&\multicolumn{2}{c}{$93.39$}
&\multicolumn{2}{c}{$97.32$}
&\multicolumn{2}{|c}{$84.11$}
\\

Top $50$
&\multicolumn{2}{c}{$56.15$} 
&\multicolumn{2}{c}{$85.31$}
&\multicolumn{2}{c}{$92.23$}
&\multicolumn{2}{c}{$98.13$} 
&\multicolumn{2}{|c}{$83.06$}
&\multicolumn{2}{c}{$93.42$}
&\multicolumn{2}{c}{$97.32$}
&\multicolumn{2}{|c}{$84.19$}
\\

{\cellcolor{myblue}Top $70$}
&\multicolumn{2}{c}{{\cellcolor{myblue}{$56.11$}}} 
&\multicolumn{2}{c}{{\cellcolor{myblue}{$85.27$}}} 
&\multicolumn{2}{c}{{\cellcolor{myblue}{$92.23$}}}
&\multicolumn{2}{c}{{\cellcolor{myblue}{$98.18$}}}
&\multicolumn{2}{|c}{{\cellcolor{myblue}{$83.40$}}} 
&\multicolumn{2}{c}{{\cellcolor{myblue}{$93.59$}}}
&\multicolumn{2}{c}{{\cellcolor{myblue}{$97.34$}}}
&\multicolumn{2}{|c}{{\cellcolor{myblue}{{$84.33$}}}}\\

Top $100$
&\multicolumn{2}{c}{$56.07$} 
&\multicolumn{2}{c}{$85.31$}
&\multicolumn{2}{c}{$92.13$}
&\multicolumn{2}{c}{$98.21$} 
&\multicolumn{2}{|c}{$83.11$}
&\multicolumn{2}{c}{$93.42$}
&\multicolumn{2}{c}{$97.32$}
&\multicolumn{2}{|c}{$84.21$}
\\

{Top $150$}
&\multicolumn{2}{c}{{{$55.94$}}} 
&\multicolumn{2}{c}{{{$85.29$}}} 
&\multicolumn{2}{c}{{{$92.20$}}}
&\multicolumn{2}{c}{{{$98.28$}}}
&\multicolumn{2}{|c}{{{$82.99$}}} 
&\multicolumn{2}{c}{{{$93.49$}}}
&\multicolumn{2}{c}{{{$97.37$}}}
&\multicolumn{2}{|c}{{{{$84.14$}}}}\\

\bottomrule
\end{tabular}

\vspace{-13pt}

\end{table*}

\begin{figure}[!t]
	\begin{center}
		\includegraphics[width=\linewidth]{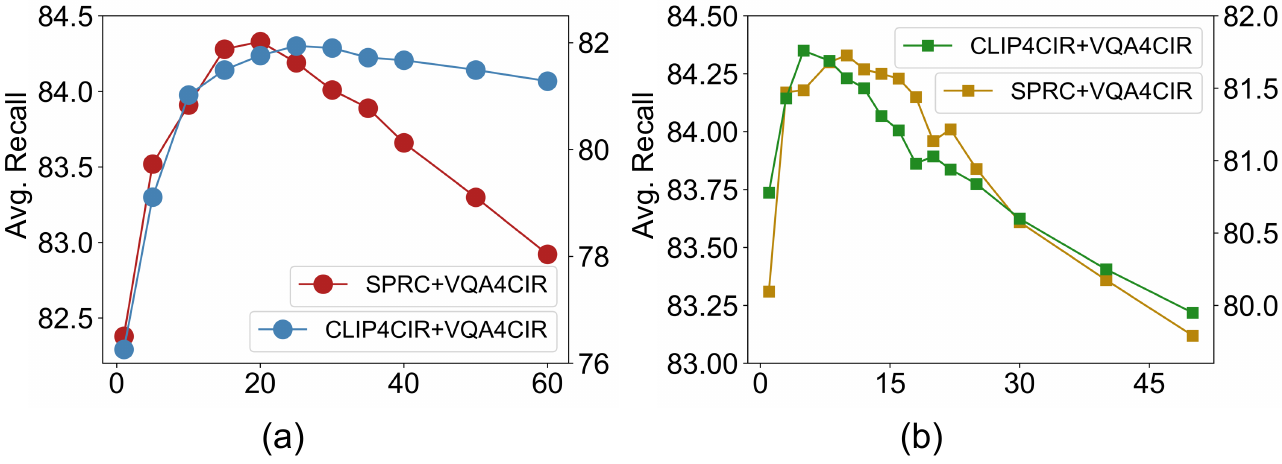}
  \put(-180,2){ \small$\alpha$}
  \put(-61,2){ \small$\beta$}
	\end{center}
    \vspace{-16pt}
	\captionsetup{font=small}
	\caption{\small\textbf{Ablation} studies of different values of \textbf{(a)} $\alpha$ and \textbf{(b)} $\beta$ on CIRR dataset.}
	\vspace{-22pt}
	\label{fig:alpha}
\end{figure}

\noindent{\textbf{Analysis of $\alpha$ and $\beta$.}}~
As mentioned in Sec.~\ref{sec:inference}, a larger $\alpha$ value indicates a greater step size in the descent, and a larger $\beta$ indicates a faster rate of decline.
%
%
Fig.~\ref{fig:alpha} shows the changes in the average recalls for CIRR under different $\alpha$ values and $\beta$ values.
As can be seen from Fig.~\ref{fig:alpha} (a), the recalls improve with an increase in $\alpha$, reaching a maximum when $\alpha$ is $25$ upon CLIP4CIR~\cite{baldrati2023composed} and $20$ upon SPRC~\cite{bai2023sentence}.
When $\alpha$ is greater than $25$, a decrease in recalls occurs. 
%
Analogously, as the $\beta$ value increases, refer to Fig.~\ref{fig:alpha} (b), the recalls on the two methods improve, reaching a maximum when $\beta$ equals $5$ and $10$, respectively.
Nonetheless, when $\beta$ becomes larger, the recalls decrease.
%

\begin{figure}[!t]
	\begin{center}
		\includegraphics[width=\linewidth]{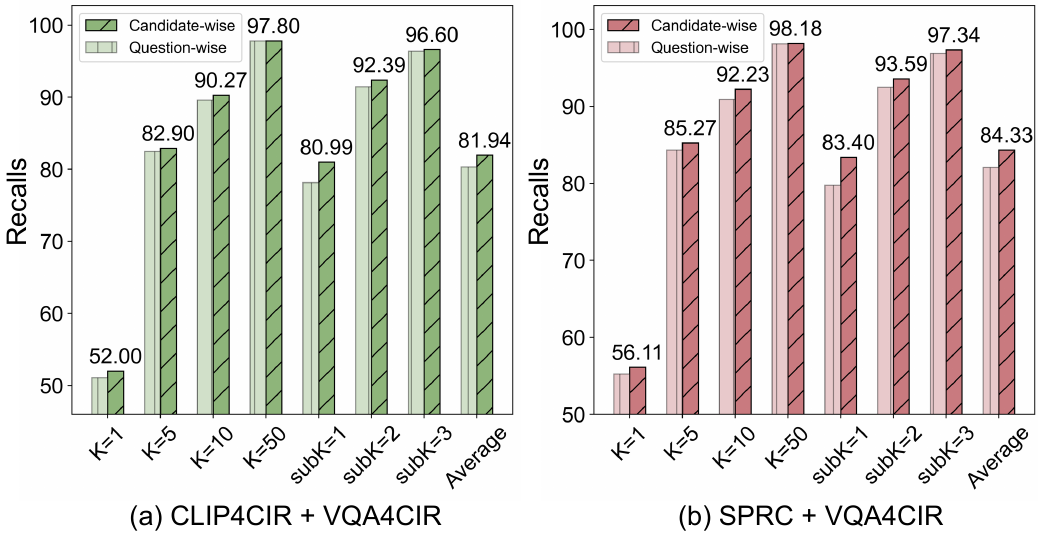}
	\end{center}
    \vspace{-16pt}
	\captionsetup{font=small}
	\caption{\small\textbf{Ablation} studies of \textit{different reranking mechanisms}, \ie, question-wise versus caption-wise on the validation set of CIRR dataset.}
	\vspace{-8pt}
	\label{fig:ReQC}
\end{figure}

\noindent{\textbf{Question-wise \textit{\textbf{vs.}} Caption-wise Re-ranking.}}~
We further discuss the effect of our method when employing a question-wise approach versus a caption-wise approach for re-ranking during the inference stage.
The caption-wise approach multiplies the prediction results for all questions to recognize whether the candidate image is consistent with the relative caption, whereas the question-wise approach re-ranks the predictions for each question directly.
We have recorded the differences in recall values between the question-wise and caption-wise approaches for the validation set of CIRR datasets of both CLIP4CIR+VQA4CIR and SPRC+VQA4CIR in Fig.~\ref{fig:ReQC}.
It can be seen that the caption-wise approach performs better than the question-wise approach because it can accurately reflect whether a candidate image is consistent with relative caption.
%
%
In light of this, we adopted the caption-wise approach for re-ranking during the inference stage in our experiments.

\begin{table*}[t]
\vspace{-13pt}
\renewcommand{\arraystretch}{1.3} 
\setlength{\tabcolsep}{15pt} 
	\caption{\small \textbf{Ablation studies} in terms of {recalls} with regard to \textit{different VQA models} on the validation set of the CIRR dataset.}
	\vspace{-8pt}
	\label{tab:llavablip}
	\fontsize{8}{8}\selectfont
	\centering
	\begin{tabular}{l cc cc cc cc cc cc cc cc}
\toprule
\multicolumn{1}{l}{\multirow{2}{*}[-1mm]{\textbf{Method}}}
&\multicolumn{8}{c}{\textbf{\texttt{Recall@K}}}
&\multicolumn{6}{c}{\textbf{\texttt{Recall$_{\text{Subset}}$}}} 
&\multicolumn{2}{c}{\multirow{2}{*}{\textbf{\texttt{Average}}}}
\\

\cmidrule(lr){2-9} \cmidrule(l){10-15}

&\multicolumn{2}{c}{\textbf{K=1}} 
&\multicolumn{2}{c}{\textbf{K=5}} 
&\multicolumn{2}{c}{\textbf{K=10}} 
&\multicolumn{2}{c}{\textbf{K=50}} 
&\multicolumn{2}{c}{\textbf{K=1}} 
&\multicolumn{2}{c}{\textbf{K=2}} 
&\multicolumn{2}{c}{\textbf{K=3}} 
&\multicolumn{2}{c}{\textbf{}} 
\\

\cmidrule(r){1-1} 
\cmidrule{2-3} 
\cmidrule(lr){4-5} 
\cmidrule(lr){6-7} 
\cmidrule(lr){8-9}
\cmidrule(lr){10-11}
\cmidrule(lr){12-13}
\cmidrule(lr){14-15}
\cmidrule(lr){16-17}

LLaVA
&\multicolumn{2}{c}{$53.53$}
&\multicolumn{2}{c}{$84.05$}
&\multicolumn{2}{c}{$91.33$}
&\multicolumn{2}{c}{$99.26$}
&\multicolumn{2}{|c}{$79.92$}
&\multicolumn{2}{c}{$93.58$}
&\multicolumn{2}{c}{$96.76$}
&\multicolumn{2}{|c}{$81.98$}

\\ 

InstructBLIP
&\multicolumn{2}{c}{$52.19$} 
&\multicolumn{2}{c}{$82.99$}
&\multicolumn{2}{c}{$91.06$}
&\multicolumn{2}{c}{$97.53$} 
&\multicolumn{2}{|c}{$78.19$}
&\multicolumn{2}{c}{$93.00$}
&\multicolumn{2}{c}{$95.99$}
&\multicolumn{2}{|c}{$80.59$}
\\

{\cellcolor{myblue}Fine-Tune LLaVA}
&\multicolumn{2}{c}{{\cellcolor{myblue}{$56.11$}}} 
&\multicolumn{2}{c}{{\cellcolor{myblue}{$85.27$}}} 
&\multicolumn{2}{c}{{\cellcolor{myblue}{$92.23$}}}
&\multicolumn{2}{c}{{\cellcolor{myblue}{$98.18$}}}
&\multicolumn{2}{|c}{{\cellcolor{myblue}{$83.40$}}} 
&\multicolumn{2}{c}{{\cellcolor{myblue}{$93.59$}}}
&\multicolumn{2}{c}{{\cellcolor{myblue}{$97.34$}}}
&\multicolumn{2}{|c}{{\cellcolor{myblue}{{$84.33$}}}}\\

\bottomrule
\end{tabular}

\end{table*}

\begin{table*}[t]
\vspace{-6pt}
\renewcommand{\arraystretch}{1.3} 
\setlength{\tabcolsep}{15pt} 
	\caption{\small \textbf{Ablation studies} in terms of {recalls} with regard to \textit{different parameter-efficient learning settings} in LLaMA~\cite{touvron2023llama} on the validation set of the CIRR dataset.}
	\vspace{-8pt}
	\label{tab:llama}
	\fontsize{8}{8}\selectfont
	\centering
	\begin{tabular}{l cc cc cc cc cc cc cc cc}
\toprule
\multicolumn{1}{l}{\multirow{2}{*}[-1mm]{\textbf{Method}}}
&\multicolumn{8}{c}{\textbf{\texttt{Recall@K}}}
&\multicolumn{6}{c}{\textbf{\texttt{Recall$_{\text{Subset}}$}}} 
&\multicolumn{2}{c}{\multirow{2}{*}{\textbf{\texttt{Average}}}}
\\

\cmidrule(lr){2-9} \cmidrule(l){10-15}

&\multicolumn{2}{c}{\textbf{K=1}} 
&\multicolumn{2}{c}{\textbf{K=5}} 
&\multicolumn{2}{c}{\textbf{K=10}} 
&\multicolumn{2}{c}{\textbf{K=50}} 
&\multicolumn{2}{c}{\textbf{K=1}} 
&\multicolumn{2}{c}{\textbf{K=2}} 
&\multicolumn{2}{c}{\textbf{K=3}} 
&\multicolumn{2}{c}{\textbf{}} 
\\

\cmidrule(r){1-1} 
\cmidrule{2-3} 
\cmidrule(lr){4-5} 
\cmidrule(lr){6-7} 
\cmidrule(lr){8-9}
\cmidrule(lr){10-11}
\cmidrule(lr){12-13}
\cmidrule(lr){14-15}
\cmidrule(lr){16-17}

Prompt
&\multicolumn{2}{c}{46.99} 
&\multicolumn{2}{c}{79.00}
&\multicolumn{2}{c}{87.99}
&\multicolumn{2}{c}{95.23} 
&\multicolumn{2}{|c}{76.99}
&\multicolumn{2}{c}{91.00}
&\multicolumn{2}{c}{94.99}
&\multicolumn{2}{|c}{77.99}
\\ 

Prompt + LoRA
&\multicolumn{2}{c}{{54.92}} 
&\multicolumn{2}{c}{{83.63}}
&\multicolumn{2}{c}{{92.24}}
&\multicolumn{2}{c}{{98.15}} 
&\multicolumn{2}{|c}{{82.07}}
&\multicolumn{2}{c}{{93.23}}
&\multicolumn{2}{c}{{97.39}}
&\multicolumn{2}{|c}{{82.85}}
\\

{\cellcolor{myblue}LoRA}
&\multicolumn{2}{c}{{\cellcolor{myblue}{$56.11$}}} 
&\multicolumn{2}{c}{{\cellcolor{myblue}{$85.27$}}} 
&\multicolumn{2}{c}{{\cellcolor{myblue}{$92.23$}}}
&\multicolumn{2}{c}{{\cellcolor{myblue}{$98.18$}}}
&\multicolumn{2}{|c}{{\cellcolor{myblue}{$83.40$}}} 
&\multicolumn{2}{c}{{\cellcolor{myblue}{$93.59$}}}
&\multicolumn{2}{c}{{\cellcolor{myblue}{$97.34$}}}
&\multicolumn{2}{|c}{{\cellcolor{myblue}{{$84.33$}}}}\\

\bottomrule
\end{tabular}

\vspace{-10pt}

\end{table*}

\begin{figure}[!t]
	\begin{center}
		\includegraphics[width=\linewidth]{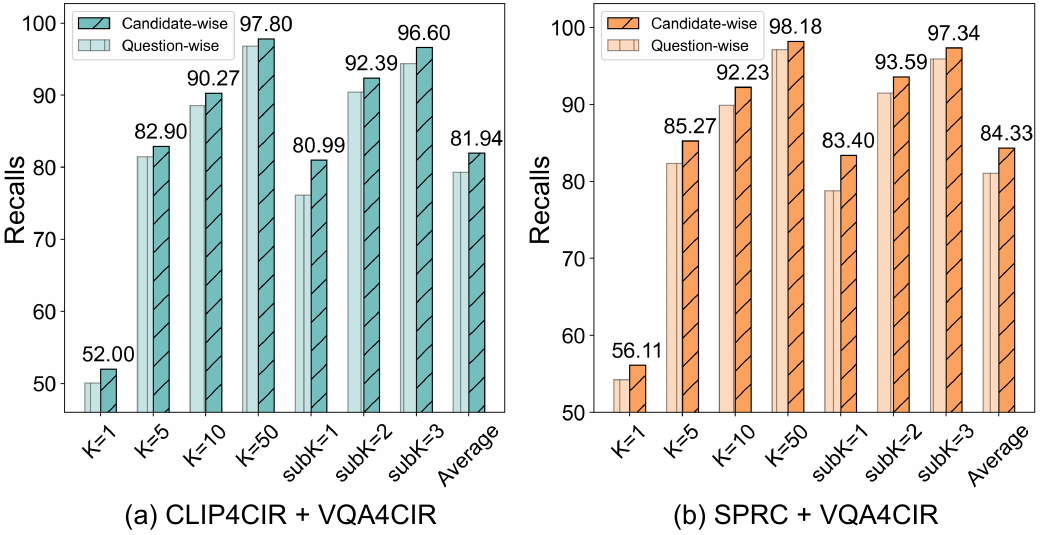}
	\end{center}
    \vspace{-16pt}
	\captionsetup{font=small}
	\caption{\small\textbf{Ablation} studies of \textit{different LLaVA fine-tune mechanisms}, \ie, question-wise versus caption-wise on the validation set of CIRR.}
	\vspace{-20pt}
	\label{fig:loss}
\end{figure}

\noindent{\textbf{Fine-tune LLaVA in Question-wise \textit{\textbf{vs.}} Caption-wise.}}~
To train LLaVA to adapt to CIR, we designed a loss to iteratively check whether the answers to all questions are correct.
Here, we discuss the differences between training LLaVA in a question-wise manner and a caption-wise manner.
We summarize the performance of these two methods on the validation sets of CIRR in Fig.~\ref{fig:loss}.
It can be seen that both methods show differences in CIRR, the average recall values of CLIP4CIR+VQA4CIR and SPRC+VQA4CIR differ by $2.62$\% and $3.27$\% respectively.
The caption-wise method provides a higher recall because it can accurately reflect whether a candidate image is consistent with relative caption.
%
%
Thus, we adopt the caption-wise method to train LLaVA in our experiments.

\noindent{\textbf{Discussion on Different VQA Models.}}~
We employ different LVLM models, \ie, the pre-trained LLaVA~\cite{liu2023visual} and instructBLIP~\cite{dai2023instructblip}, as VQA models to assess the effect of  LVLM.
We provide the performance of the two variants on the validation sets of two datasets in Table~\ref{tab:llavablip}.
One can see that both approaches significantly underperform our fine-tuned LLaVA~\cite{liu2023visual} because directly using pre-trained LVLMs does not generalize well to our task.
Moreover, using the pre-trained LLaVA~\cite{liu2023visual} and instructBLIP directly do not show a substantial difference in recall values, \ie, average recalls: LLaVA~\cite{liu2023visual}: $81.98$ \textit{\textbf{vs.}} InstructBLIP~\cite{dai2023instructblip}: $80.59$.
In contrast, our method, which only fine-tunes a small number of parameters with LoRA can achieve noticeably improved performance, with average recalls: $81.98$ $\&$ $80.59$ {\textit{vs.}} $\textbf{84.33}$.

\noindent{\textbf{Learning of LLaMA Models.}}~
In Sec.~\ref{sec:llama}, we mentioned that during the training of LLaMA, we fixed the prompt and only learned a small number of parameters with LoRA.
Here, we discuss alternative learning methods, \ie, learning both the prompt and LoRA simultaneously and learning only the prompt.
We provide the results of different learning mechanisms on the CIRR validation set in Table~\ref{tab:llama}.
From the table, the method of fixing the prompt is superior to the method of learning both the prompt and LoRA simultaneously, \ie, Avg.: $\textbf{84.33}$ \textit{{vs.}} $82.85$.
The method of learning only the prompt performed the worst.
%
%
Given the above results, we adopt the method of fixing the prompt and only train LoRA, where the rank size of LoRA is empirically set to $8$.

%% file: sec/5_Conclusion.tex
\section{Conclusion}
This paper provides a VQA perspective for boosting CIR performance and suggest the VQA4CIR method.
Our VQA4CIR is built upon the fact that a certain percentage of failure retrieval results in existing CIR methods are not consistent with their relative captions.
To find the inconsistent images, we adopted the "QA generation → VQA" self-verification pipeline. 
First, we leverage fine-tuned LLaMA to generate QA pairs from relative caption.
Then, fine-tuned LLaVA was adopted as the VQA model for finding the images inconsistent with relative caption.
The inconsistent images are then reranked to enhance CIR performance.
Our VQA4CIR can be incorporated with most existing CIR methods. 
Experiments show that our VQA4CIR outperforms the state-of-the-art methods on the CIRR and Fashion-IQ datasets. 
Nonetheless, our VQA4CIR relies on LLM and LVLM for re-ranking has a higher cost at inference time.
%
%
In the future, we will investigate a more efficient solution for verifying the consistency between retrieved images and relative captions.

%
%
%
%
%